\documentclass[preprint,12pt]{article}  

\usepackage{graphicx}
\graphicspath{{./figures/}{./inkscape/}{../figures/}} 
\usepackage{amsmath,amsfonts,amsthm,bm,mathtools} 
\usepackage{amssymb} 
\usepackage{lipsum}
\usepackage{float}
\usepackage{pgfplots}
\usepackage{graphicx,amsmath,upgreek,bm}
\usepackage{subcaption}
\usepackage{siunitx}
\usepackage{makecell}
\usepackage{booktabs}
\pgfplotsset{compat=newest}
\usepackage{multicol}
\usepackage{xcolor,colortbl}
\pgfplotsset{plot coordinates/math parser=false}
\usepackage{multicol} 
\usepackage{amsthm}

 \usepackage[numbers]{natbib}  
\usepackage{geometry}
\usepackage{afterpage}
\usepackage{longtable}
\usepackage[tableposition=top]{caption}
\usepackage{multirow}
\usepackage[bb=boondox]{mathalpha}
\definecolor{label}{rgb}{0.1,0.25,.65}
\definecolor{title}{rgb}{0.1,0.25,.65}

\definecolor{label}{rgb}{0,0,0}
\definecolor{title}{rgb}{0,0,0}

\definecolor{cite}{rgb}{0.6,0.1,.2}
\usepackage[
	font=normalsize,
	labelfont={rm,bf,color=label}, 
	justification=justified,
	format=plain]{caption} 
\usepackage{subcaption}
\newlength\fwidth
\usepackage{natbib}
\usepackage{bibentry}
\definecolor{best_acc}{rgb}{0,0.5,0}
\definecolor{refcol}{rgb}{.25,0,1}
\definecolor{mygray}{gray}{0.85}
\definecolor{mycite}{gray}{0.0}
\let\oldbibliography\thebibliography
\renewcommand{\thebibliography}[1]{%
	\oldbibliography{#1}%
	\setlength{\itemsep}{0pt}%
}
\definecolor{mygray}{gray}{0.85}
\definecolor{edit1}{rgb}{.375,.375,.375}
\definecolor{edit2}{rgb}{1,0,0}

\usepackage[misc]{ifsym}
\usepackage{fontawesome}
\usepackage{ragged2e}
\title{
	\Large{\textbf{Probabilistic Emissivity Retrieval from Hyperspectral Data via Physics-Guided Variational Inference}}
}
\author{
\small	Joshua R.~Tempelman\textsuperscript{1}\thanks{Corresponding author: tempelman@lanl.gov} \and
\small	Kevin Mitchell\textsuperscript{2} \and
\small	Adam J.~Wachtor\textsuperscript{3}  \and
\small	Eric B.~Flynn\textsuperscript{1}
}
\date{\footnotesize
	\textsuperscript{1} Space Remote Sensing and Data Science, Los Alamos National Laboratory, Los Alamos, NM, USA\\
	\footnotesize
	\textsuperscript{2} Physical Chemistry and Applied Spectroscopy, Los Alamos National Laboratory, Los Alamos, NM, USA\\
	\footnotesize
	\textsuperscript{3} Engineering Institute Los Alamos National Laboratory, Los Alamos, NM, USA
}

\usepackage{titlesec}
\titleformat{\section}
{
	\bfseries
	\large
	\color{label}
}  
{\thesection}  
{1em}  
{}

\titleformat{\subsection}
{
	\bfseries
	\normalsize
	\color{label}
}  
{\thesubsection}
{1em}
{}

\titleformat{\subsubsection}
{
	\bfseries
	\itshape
	\normalsize
	\color{label}
}  
{\thesubsubsection}
{1em}
{}

\usepackage{todonotes}
\usepackage{lineno}

\usepackage{setspace}
\newcommand{\fullfig}[1]{\includegraphics[width=.5\linewidth]{#1}}
\newcommand{\fullfigdouble}[1]{\includegraphics[width=\linewidth]{#1}}
\newcommand{\subfigurewidth}{.25\linewidth}
\newcommand{\appendices}{\appendix}

\usepackage{mathrsfs}
 
\usepackage[colorlinks=true,
linkcolor=cite,
citecolor=cite,
urlcolor=cite]{hyperref} 
\usepackage{tcolorbox}

\newcommand{\makeabstract}[2]{%
	\small \noindent\textbf{#1}~{\small #2}
}

\begin{document}
	
	\maketitle

	\makeabstract{Abstract:}{
		Recent research has proven neural networks to be a powerful tool for performing hyperspectral imaging (HSI) target identification.
		However, many deep learning frameworks deliver a single material class prediction and operate on a per-pixel basis; such approaches are limited in their interpretability and restricted to predicting materials that are accessible in available training libraries.  
		In this work, we present an inverse modeling approach in the form of a physics-conditioned generative model.
		A probabilistic latent-variable model learns the underlying distribution of HSI radiance measurements and produces the conditional distribution of the emissivity spectrum. Moreover, estimates of the HSI scene’s atmosphere and background are used as a physically relevant conditioning mechanism to contextualize a given radiance measurement during the encoding and decoding processes.  Furthermore, we employ an in-the-loop augmentation scheme and physics-based loss criteria to avoid bias towards a predefined training material set and to encourage the model to learn physically consistent inverse mappings. Monte-Carlo sampling of the model's conditioned posterior delivers a sought emissivity distribution and allows for interpretable
		 uncertainty quantification.
		Moreover, a distribution-based material matching scheme is presented to return a set of likely material matches for an inferred emissivity distribution.
		Hence, we present a strategy to incorporate contextual information about a given HSI scene, capture the possible variation of underlying material spectra, and provide interpretable probability measures of a candidate material accounting for given remotely-sensed radiance measurement.
	}\\
	
	 \makeabstract{Keywords:}{Remote Sensing, Hyperspectral Imaging, Emissivity Retrieval, Variational Modeling}
		  
	\vspace*{-0em} 
	\begin{center}
		\begin{tcolorbox}[colback=white, colframe=gray, width=\textwidth]
			\small   \textit{This work has been submitted to the IEEE for possible publication. Copyright may be transferred without notice, after which this version may no longer be accessible. }
		\end{tcolorbox}
	\end{center}
	\newpage
	
	


%
\section{Introduction} 
\label{SEC:Intro}
 
Long-wave infrared (LWIR) hyperspectral imaging (HSI) is an effective means for identifying materials from remote (often aerial) sensors thanks to the unique atmospheric and thermal emission properties in the 7-to-14 $\mu$m regime of the electromagnetic spectrum~\cite{Manolakis2002,Manolakis2019}. 
As such, common application spaces of LWIR-HSI technologies include agricultural production monitoring~\cite{Dale2013,Garcia2024}, mineral exploration~\cite{Sabins1999}, and urban monitoring~\cite{heiden2012urban,ramdani2013urban}.
LWIR-HSI frameworks for target material identification (ID) may generally be split into two categories: radiance-based classifiers  and  emissivity-retrieval models.
The former seeks to draw either a categorical or binary decision regarding the ground-based material given the observed radiance~\cite{zhang2020htd,zhu2020two}, while the latter seeks to first estimate the spectral emissivity of the surface, which may be subsequently  matched to those of  known substances~\cite{Boonmee2006,Li2007,Xu2021,Mcelhinney2022}.
While material ID models are convenient in the sense that their decision making is largely performed under-the-hood, their interpretation is not straight-forward, as they are typically (i) restricted to providing categorical answers with limited clarity regarding uncertainty quantification, and (ii) can only predict materials that are known a-priori to the model, e.g., training materials~\cite{Sifnaios2024,Klein2023}.
Likewise, while numerous atmospheric compensation and temperature-emissivity decoupling techniques exist to perform emissivity retrieval~\cite{Buckland2017}, they suffer from the cumulative build-up of uncertainty in solving an ill-posed and over-parameterized inverse problem~\cite{Distasio2010}.
Hence, despite the promise of HSI technologies~\cite{Khan2018}, developing interpretable imagining methods still presents multiple challenges due to environmental and atmospheric variation which occlude LWIR-HSI measurements, hence making the task of atmospheric and background estimation a vital component of practical HSI implementation~\cite{Acito2019}.

In recent works, LWIR-HSI frameworks have been supplemented by the introduction of machine learning models which have proven to be well-suited for the task of material ID~\cite{Signoroni2019,Rasti2020,Gupta2022}.
Moreover, the flexibility offered by deep-learning architectures and training schemes have allowed for the easy inclusion of multi-modal information, such as LiDar~\cite{Xue2021,Tan2024}, as well as physics-based guidance \cite{Klein2023}.
As such, many current HSI efforts utilize deep-learning methodologies such as fully-connected and graph neural networks to conduct HSI exploitation~\cite{Wang2021}.
However, despite the flexibility of modern deep-learning based approaches in learning complex recognition tasks, they are generally limited in their interpretability and are commonly applied in a black-box manner. 

While recent work has shed light on spectral feature attribution in the context of material classifiers~\cite{Klein2023a}, the capacity to account for uncertainty in the over-parameterized inversion of the propagation physics and to provide a candidate set of likely spectral matches to a ground material remains an elusive task.
For this reason, we seek a deep-learning based emissivity retrieval model, e.g., an  \textit{inverse model} that reverses the radiance propagation process and returns the spectra of a target material in a (possibly) mixed pixel.
To robustly configure such a retrieval model, several challenges must be addressed.
Namely, stochastically varying atmospheric conditions distort spectral signatures, requiring the compensation of atmospheric effects	\cite{Guanter2007,Distasio2010,Thompson2016}.
Additionally, pixels may be composed of a multiple mixed materials, with some being of interest and others not, leading to further complications in recovering a target spectrum~\cite{Patel2020}.
Moreover, the inverse problem is ill-posed, making the process of correctly decoupling the effects of atmospheric propagation from the material emissivity highly complicated.
For this reason, it's desirable to devise a more flexible modeling framework that is capable of incorporating uncertainty in the inverse process to provide distribution-based estimates of the retrieved material emissivity over the LWIR spectrum.

In this work, we address these challenges by considering a probabilistic inverse model that is conditioned on the outputs maxillary of models which estimate the scene's radiance propagation parameters the scene-wide background (non-target).
We focus explicitly on surface emissivity retrieval of solid materials (e.g., gas plumes are not considered).
Whereas prior works have investigated probabilistic modeling for estimating the uncertainty of class-label predictions~\cite{He2023} and image segment classification~\cite{Seydgar2022}, this work is concerned with learning the aleatoric uncertainty of the inverse problem itself.
Moreover, the  modeling scheme proposed herein focuses on incorporating knowledge of the radiance propagation process.
While recent work has demonstrated the auxiliary measurements from an HSI cube are beneficial for contextualizing a pixel of interest, commonly via graph neural networks~\cite{Yao2023,Ding2024,Wu2024}, this work differs by focusing explicitly on learning the radiance propagation properties of HSI scenes to mitigate the over-parameterization of the inverse process.
Our model takes the form of a conditioned variational latent variable model with auxiliary networks that produce conditional atmosphere and background estimates.
In this light, our model is \textit{physics-guided}, since it utilizes knowledge of the propagation physics to contextualize the radiance measurements of a given pixel in light of the inversion task.

Accordingly, the remainder of this paper is organized as follows. Section~\ref{SEC:HSI} reviews the relevant fundamentals of HSI radiance propagation and provides the mathematical framework for the proposed probabilistic inverse model.
Section~\ref{SEC:ML} outlines the machine learning architecture and information regarding training objectives and learning schemes.
The results of several numerical studies are presented in Section~\ref{SEC:Results} which highlights the model's performance on hold-out HSI data.
Lastly, Section~\ref{SEC:Conclusions} offers concluding remarks and suggestions for future research.

%


\ifdefined\MYCOMMANDS

\else
    \def\MYCOMMANDS{} 

    \newcommand{\eps}{{\epsilon}}
    \newcommand{\rad}{{L}}
    \newcommand{\radw}{\rad_w}
    \newcommand{\atm}{\mathcal{A}}
    \newcommand{\z}{{z}}
    \newcommand{\radset}{\underline{L}}

    \newcommand{\trans}{{\tau}}
    \newcommand{\bbrad}{{B}}
    \newcommand{\Lu}{\rad_u}
    \newcommand{\Ld}{\rad_d}
    \newcommand{\Lt}{\rad_{\text{t}}}
    \newcommand{\Lbg}{\rad_{\text{bg}}}

    \newcommand{\cube}{\mathsf{C}}
    \newcommand{\cubeset}{[\cube]}
    \newcommand{\atmset}{[\atm]}
    \newcommand{\bgset}{[\Lbg]}
    \newcommand{\epsset}{[{\eps}]}

    \newcommand{\mlp}{\text{mlp}}
    \newcommand{\fno}{\mathcal{U}}
    \newcommand{\dec}{{\mathsf{D}}}
    \newcommand{\enc}{{\mathsf{E}}}
    \newcommand{\flow}{f_z}
    \newcommand{\encodervae}{x_{\enc_{\rad}}}
    \newcommand{\decodervae}{x_{\dec_{\rad}}}
    \newcommand{\epsnetvae}{x_{\dec_{\eps}}}
    \newcommand{\epslayer}{x_{\texttt{e}}}

    \newcommand{\epsfspace}{\mathscr{E}}
    \newcommand{\radfspace}{\mathscr{L}}
    \newcommand{\zfspace}{\mathscr{Z}} 

    \newcommand{\Lbgmean}{\bar{L}_{\text{bg}}^\cube}

    \newcommand{\Lbgmeanhat}{\hat{\bar{L}}_{\text{bg}}^{\cube}}
    \newcommand{\Lbgmeank}{\bar{L}_{\text{bg}}^{\cube_k}}
    \newcommand{\Lbgmeankhat}{\hat{\bar{L}}_{\text{bg}}^{\cube_k}}


    \newcommand{\LL}{ \ell{(\eps|\rad)} }
    \newcommand{\ELBO}{ \text{ELBO}_{(\eps|\rad)} }
    \newcommand{\kld}{D_{\text{KL}}}
    \newcommand{\iter}{\jmath}
    \newcommand{\ldim}{d_z}
    \newcommand{\cdim}{d_c}
    \newcommand{\bgdim}{d_{\text{bg}}}
    \newcommand{\atmdim}{d_{\text{prop}}}
    \newcommand{\concat}{\oplus}
    \newcommand{\concatc}{ : }

    \newcommand{\atmnet}{\texttt{PropNet}}
    \newcommand{\bgnet}{\texttt{BgNet}}
    \newcommand{\epsnet}{\texttt{EpsNet}}
\fi

\section{Hyperspectral Inverse Problem and Mathematical Framework}
\label{SEC:HSI}
We tailor our modeling around the classical radiance propagation equation for LWIR-HSI modeling~\cite{Manolakis2019}. This section serves to review its formulation and the classical whitening transformation, motivate the need for a probabilistic modeling framework to learn its inverse, and present a mathematical framework to do so.

\subsection{Radiance Propagation Model and Data Pre-processing}

\begin{figure}[t!]
	\begin{subfigure}{\linewidth}\centering
		\fullfig{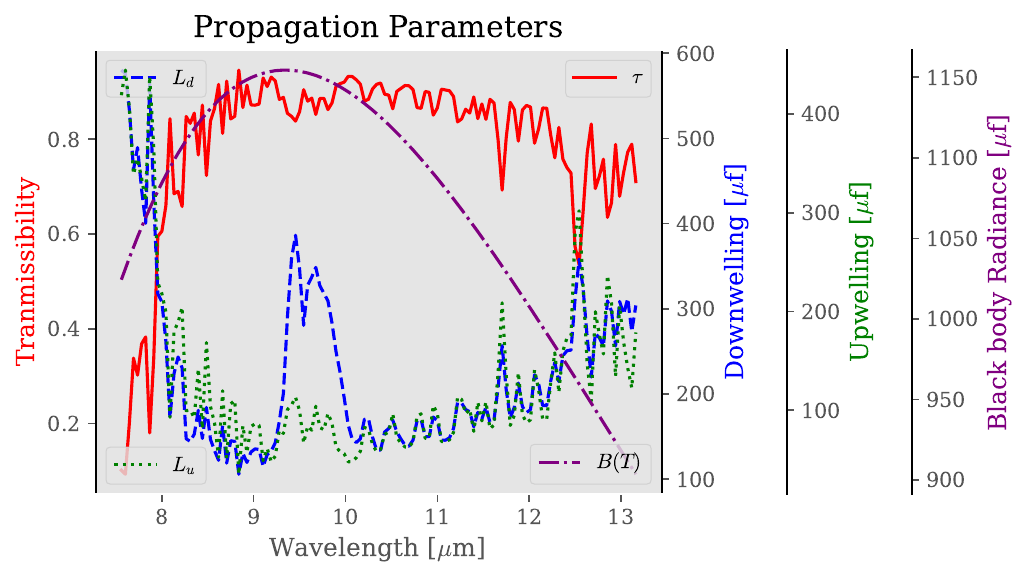}
		\caption{Example of the atmospheric transmission variables $\atm$, depicting transmissibility (left-axis), downwelling (inside-right axis), upwelling (middle-right axis), and black-body radiance (far-right axis).}
	\end{subfigure}
	\begin{subfigure}{\linewidth}\centering
		\fullfig{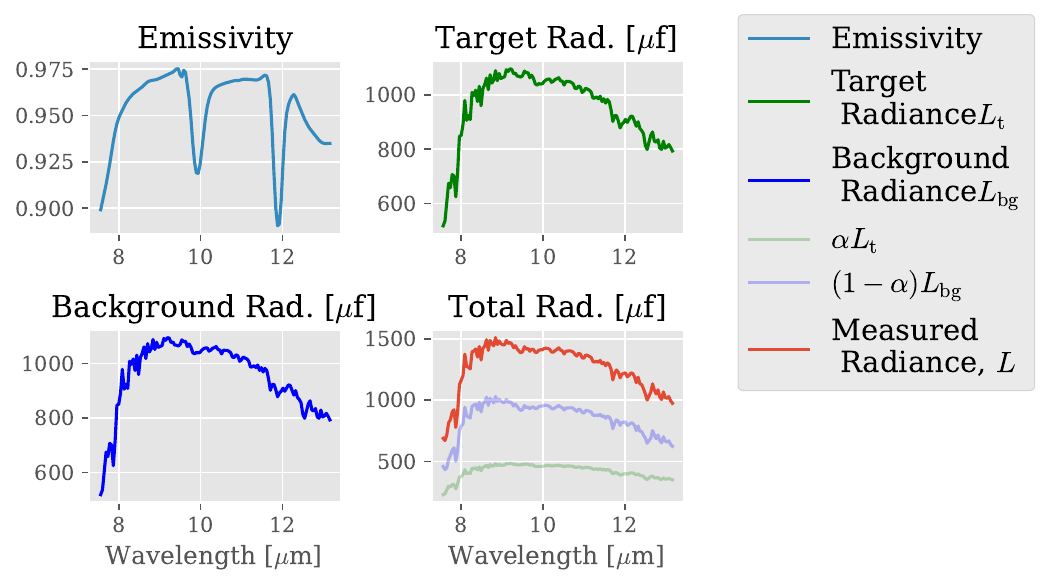}
		\caption{Example emissivity spectra $\eps$, background radiance $\Lbg$, and propagated radiance $\rad=f_{\text{p}}(\eps;\atm,\Lbg)$ per the atmospheric parameters of subplot (a).}
	\end{subfigure}
	\caption{Example of propagated radiance and atmospheric parameters. (a) The atmospheric parameters $\atm$ and (b) the target emissivity $\epsilon$, pixel background radiance $\Lbg$, target radiance $\Lt$ and propagated radiance $\rad$.}
	\label{Fig:Lp}
\end{figure}
%
The radiative transfer of thermal emissions to an airborne sensor is a function of the material emissivity $\epsilon$, downwelling radiance $L_d$, upwelling radiance $\Lu$, transmission $\tau$, and the temperature-dependent black-body radiance $B(T)$ with $T$ being the average temperature of the scene.
In this work, we assume the material reflectance is accurately approximated by $\rho=1-\epsilon$.
The two primary components accounting for the measured radiance are ground-leaving radiance and ambient radiance.
The first is a combination of the ground material emission $\eps\bbrad(T)$ and the reflected radiance  $\Ld\rho$, while the latter is comprised of   $L_u$.
The atmospheric transmission, $\tau$, determines how much of the ground-leaving radiance reaches the sensor as a function of wavelength~\cite{Manolakis2019}.
In practice, a material of interest may not fully cover a pixel of an HSI measurement, and hence we consider mixed pixels of target and background radiance of a given pixel (denoted $\Lbg$)  via the relative strength factor $\alpha$.
Combining target and background radiance, the total at-sensor radiance, $\rad=f_{\rm p}(\eps; \mathcal{A}, \alpha)$, is calculated as,
\begin{equation}
		f_{\rm p}(\eps; \mathcal{A}, \alpha) = \alpha\left[
		\underbrace{\trans (\eps \bbrad(T) +  (1-\eps  )\Ld ) + \Lu}_{\Lt (\eps,\atm)}
		\right]
		+ (1-\alpha)\Lbg
	\label{eq:fp}
\end{equation}
where $\Lt$ is the radiance propagated from the target material to the sensor and  $\mathcal{A} = \{\bbrad(T),\trans,\Lu,\Ld\}$ the set of radiance propagation parameters, e.g., the temperature-dependent Plank function and atmospheric parameters which together are responsible for $\Lt$ and $\Lbg$. The units for wavelength and radiance are given in microns [$\mu$m] and microflicks [$\mu$f] unless otherwise stated. Fig.~\ref{Fig:Lp} provides a visualization of the various components germane to Eq~\eqref{eq:fp}.

\begin{figure}[t!]
	\centering
	\fullfig{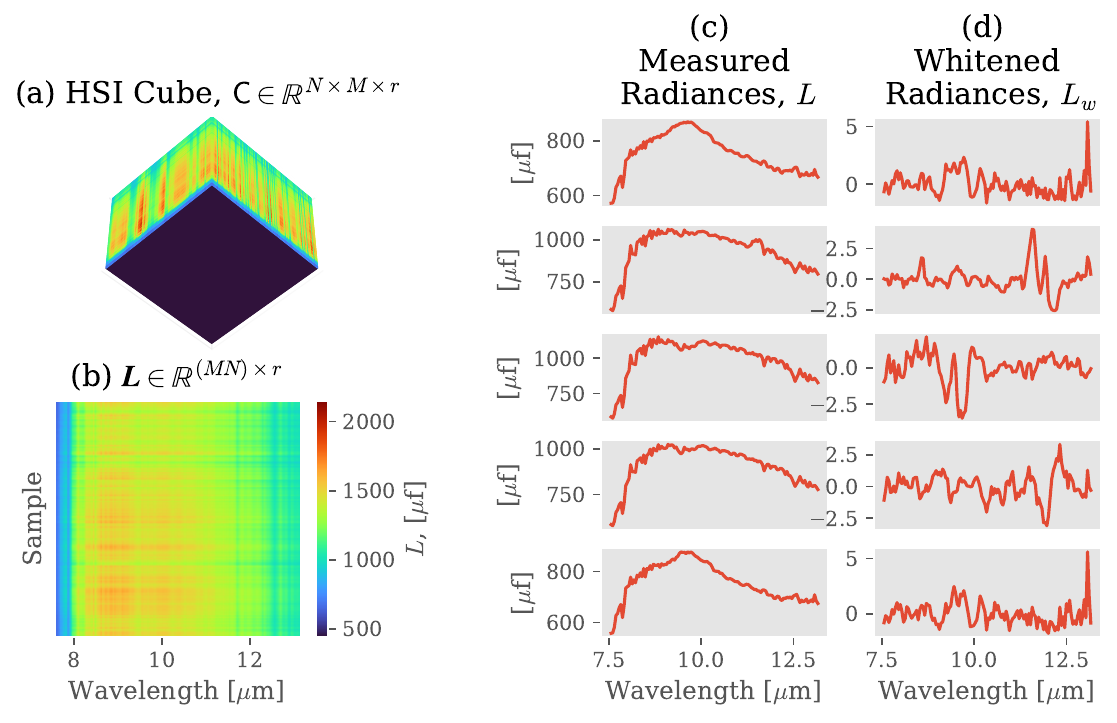}
	\caption{The cube-whitening process depicting (a)  the measured  hyperspectral data cube  $\mathsf{C}$ and (b) its reshaped matrix form $\bm\rad = \cube_{(1,2)}\in\mathbb{R}^{(N M)\times r}$ accompanied by specific examples of (c) un-whitened and (d) whitened radiance measures. }
	\label{Fig:Lw}
\end{figure}

Collecting radiance observations over a scene provides a cube of HSI data, $\cube\in\mathbb{R}^{N\times M\times r}$ where $N$, $M$, and $r$ denote the number of width-wide pixels,  height-wise pixels, and the sensor resolution, respectively, where $M=r=128$ and $N$ varies per cube.
Due to the colloquial assumption that $\atm$ and $\Lbg$ are generally consistent throughout a given HSI scene~\cite{Manolakis2019,Manolakis2016,Gao2006}, there is typically strong correlation between radiance measures of a given cube. As such, the matrix form of a cube's radiance measures $\bm\rad = \cube_{(1,2)}\in\mathbb{R}^{(N M)\times r}$ may be decoupled via a whitening transformation,  
\begin{equation}
	\bm{\rad}_w = (\bm{\rad}-\Lbgmean)\textbf{W}, 
\end{equation}
where $\Lbgmean$ is the outlier removed mean background radiance, $\textbf{W}=\textbf{V}\bm{\Lambda}^{-\frac{1}{2}}\textbf{V}^{\intercal}\in\mathbb{R}^{r\times r}$ is the whitening matrix with $\bm{\Lambda}\in\mathbb{R}^{r\times r}$ and $\textbf{V}\in\mathbb{R}^{r\times r}$ being products of the eigendecomposition of the covariance, $\bm{\Sigma}=\textbf{V}\bm{\Lambda}\textbf{V}^\intercal$.
In the current work, the covariance is estimated with a subset of $\bm{\rad}$, with  20\% of the pixels being removed as outliers.
Accordingly, for each pixel in the collection $\cube$, the set $\radset = \{{\rad},\rad_w\}$ is returned. Fig~\ref{Fig:Lw} depicts the correlated radiance measures of an example cube $\cube$, its reshaped matrix form $\bm{L}$, and several  examples of unwhitened and whitened measurements.
Because a majority of the correlation in $\bm{\rad}$ is a product of $\atm$ and $\Lbgmean$, the whitened signals provide a more expressive representation of the pixel-to-pixel variation which emphasizes specific spectral features of $\Lt$ as compared to ${\rad}$ alone. 
In reality, $\Lbgmean$ is not constant throughout the scene, so the pixel-wise background variation of $\Lbg$ is also present in $\rad_w$.
Moreover, weakly present signatures (e.g., with a small $\alpha$ value), results in less prominent signatures in $\rad_w$. 

\subsection{Motivation and Modeling Goal}
\label{subsec:motivation}
The task of emissivity retrieval is traditionally performed via atmospheric compensation and temperature-emissivity separation~\cite{Manolakis2019}.
This is commonly achieved by  estimating a scene's atmospheric make-up, i.e., by regressing over a sequence of candidate templates~\cite{Adler2014} or empirically estimating in-scene atmospheres based on statistical properties of the cube~\cite{Young2002}. 
While doing so enables the estimation of the underlying emissivity of a scene, such approaches are limited in the sense that they cannot accommodate all possible variations in the atmosphere and background, particularly since many softwares reduce the problem to a low-parameter least-squares matching with predefined atmospheric templates.
Moreover, the sought inverse is over-parameterized and ill-posed, rendering it numerically challenging.
To this end, recent works have demonstrated that knowledge of the propagation equation, coupled with estimates of the scene's atmospheric composition through numerical optimization programs, allows for the training of \textit{physics-guided} deep learning frameworks which learn to reconstruct the radiance via Eq~\eqref{eq:fp} by inferring all parameters given a single radiance measure~\cite{Klein2023}.
However, the aforementioned framework is tuned to deterministically perform on a subset of defined atmospheric templates and known material spectra, leading to inherent bias and inability to account for
uncertainty in the radiance propagation process brought-about by un-modeled factors such as turbulence or atmospheric heterogeneity~\cite{Rahm2023}, or to account for spectral signals not known to the training library.

For the above reasons, we seek a probabilistic model that, along with estimating an expected emissivity of a given pixel of an HSI scene, incorporates the flexibility of probabilistic modeling and learns the underlying uncertainty of the   ill-posed inverse problem. 
To this end, we seek to learn the probabilistic mapping $f$ based on labeled pairs of data $(\rad,\eps)$, with $\rad \in \radfspace$ being observed radiance $\eps \in \epsfspace$ the underlying target emissivity, which is also conditioned on estimates of the HSI scenes atmosphere and background (denoted $\hat\cube$), e.g., 
\begin{equation}
	f(\rad|\hat\cube):   \radfspace   \to \epsfspace  , \ \ \ f(\rad|\hat\cube):\mathbb{R}\to\mathbb{R},
\end{equation}
such that wavelength-dependent uncertainty may be captured.
To this end, our mathematical objective is to infer the marginal conditional distribution $p(\eps|\rad)$.

\subsection{Mathematical Framework}

We begin with the assumption that the emissivity and radiance are related via a
joint distribution $p(\rad,\eps,\z): \radfspace \times \epsfspace \times \zfspace\to\mathbb{R}_{+}$, where $\z \in\zfspace$ is a latent variable.
Conditional independence between $\rad$ and $\eps$ is assumed through the latent variable $\z$, such that, 
\begin{equation} 
p(\rad,\eps,\z)=p(\rad)p(\z|\rad)p(\eps|\z),
\label{Eq:p_factor}
\end{equation}
meaning that the shared-uncertainty of the inverse mapping may be proxied by an underlying posterior latent distribution that is to be learned.
The conditional independence assumption between $\rad$ and $\eps$ given $\z$ allows the latent space to serve as a manifold where all input-output pairings collapse onto structured representations,
allowing the model to learn meaningful latent representations while maintaining a probabilistic structure.

The mathematical objective of the proposed framework is to maximize   the log-likelihood of $p(\eps|\rad)$:
\begin{equation}
	\begin{aligned}
		\max \LL& = \log p(\eps|\rad) \\
		& = \log \int p(\eps,\z|\rad){\rm d}\z.
	\end{aligned} 
\end{equation}
By Eq~\eqref{Eq:p_factor}, this may be equivalently expressed as,
\begin{equation}
		\LL =\log\int p(\z|\rad)p(\eps|z){\rm d}\z.
	\label{EQ: log_liklihood_p(y|x)}
\end{equation}
Eq~\eqref{EQ: log_liklihood_p(y|x)} is intractable due to the integration over the latent variable. To account for this, a variational approximation of the latent posterior $q(\z|\rad)$ is introduced so that the  objective becomes:
\begin{equation}
	\begin{aligned}
		\LL 
		&=\log \mathbb{E}_{\z\sim q(\z|\rad)}\left[\frac{p(\z|\rad)p(\eps|\z)}{q(\z|\rad)}\right]\\
		& \geq  \mathbb{E}_{\z\sim q(\z|\rad)}\left[\log \frac{p(\z|\rad)p(\eps|\z)}{q(\z|\rad)}\right]
	\end{aligned}
	\label{EQ: log_liklihood_expectation}
\end{equation}
with the lower-bound on Eq~\eqref{EQ: log_liklihood_expectation} being a result of Jensen's inequality, thus making it the evidence lower bound (ELBO) of $\LL$ (denoted $\text{ELBO}_{\eps|\rad}$).
Applying Bayes theorem to $p(z|\rad$) and factoring out distributions constant under the expectation, the ELBO objective may be written explicitly as:
\begin{equation}
	\begin{aligned}
		\text{ELBO}_{\eps|\rad}
		=& \mathbb{E}_{z\sim q(z|\rad)}\left[
		\log p(\rad|z) + \log p(\eps|z)\right]\\& - \kld\left[q(z|x)||p(z)\right]
	\end{aligned}
	\label{EQ:ELBO}
\end{equation}
where $ \kld\left[q(z|\rad)||p(z)\right]$ is the Kullback–Leibler (KL) divergence between $q(z|\rad)$ and $p(z)$, where $p(z)=\mathcal{N}(0,\mathbf{I})$ herein.
Eq~\eqref{EQ:ELBO} may be evaluated via amortized inference by considering heteroscedastic Gaussian distributions for $q_\phi(z|x)$, $p_\theta(\rad|z)$ and $p_\gamma(\eps|z)$, where $\phi$, $\theta$, and $\gamma$ are the sought parameters of the assumed model forms.
A complete derivation of Eq~\eqref{EQ:ELBO}, as well as its numerical implementation, is given in~\ref{Apx:ELBO}. 

In addition to the evidence $\rad$, we  apply a conditioning mechanism in the form of the HSI cube's estimated atmosphere and background (see section~\ref{subsec:conditioning} for details).
The set of conditioning vectors applied to the input and latent spaces of the generative model are denoted as $\hat{\mathsf{C}}  = \{ \hat{\mathcal{A}},\Lbgmeanhat\}$ and $c=\{c_{\rm prop},c_{\rm bg}\}$, respectively, so that the ELBO loss which is optimized in the learning process is more completely written as:
\begin{equation}
	\begin{aligned}
		\text{ELBO}_{\eps|\rad}
		=& \mathbb{E}_{z\sim q_\phi(z|\rad)}\left[
		\log p_\theta(\rad|z,c) + \log p_\gamma(\eps|z,c)\right] \\ &-\kld\left[q_\phi(z|x,\hat{\mathsf{C}})||p(z)\right]
\end{aligned}
\label{EQ:ELBO_c}
\end{equation}
In this scheme, $\hat{\mathsf{C}}$ estimates the five unknown variables of Eq~\eqref{eq:fp} (excluding the inversion target  $\eps$), while $c$ serves as a latent representation of $\hat{\mathsf{C}}$. 

\section{Machine Learning Framework}
\label{SEC:ML}

The machine learning framework follows the conventional supervised learning paradigm.
We consider a collection of $K$ HSI cubes recovered from airborne measurements, $\cubeset= \{\mathsf{C}_k\}_{k=1}^{K}$, as well as a collection of $M$ laboratory-measured spectral signatures, $\epsset  = \{\epsilon_m\}_{m=1}^{M}$. 
For each of the $K$ cubes, the $M$ signatures were injected pixel-wise into the HSI scene per Eq~\eqref{eq:fp} with varying strengths, with the parameters $\atm$ coming from conventional physics-based atmospheric estimation routines~\cite{Manolakis2002}, and the pixel background $\Lbg$ being taken directly from the measured HSI scene.
Further details regarding the particulars of the constructed data set may be found in section~\ref{subsec:dataset}.
The result is a collection of labeled data pairs $(\rad,\eps)$, with accompanying cube parameters $\atm$ and $\Lbgmean$, which may be leveraged to learn the distribution $p(\epsilon|L)$.

\subsection{Data Flow and Monte-Carlo Approximation}

\begin{figure*}[t!]
	\centering
	\fullfigdouble{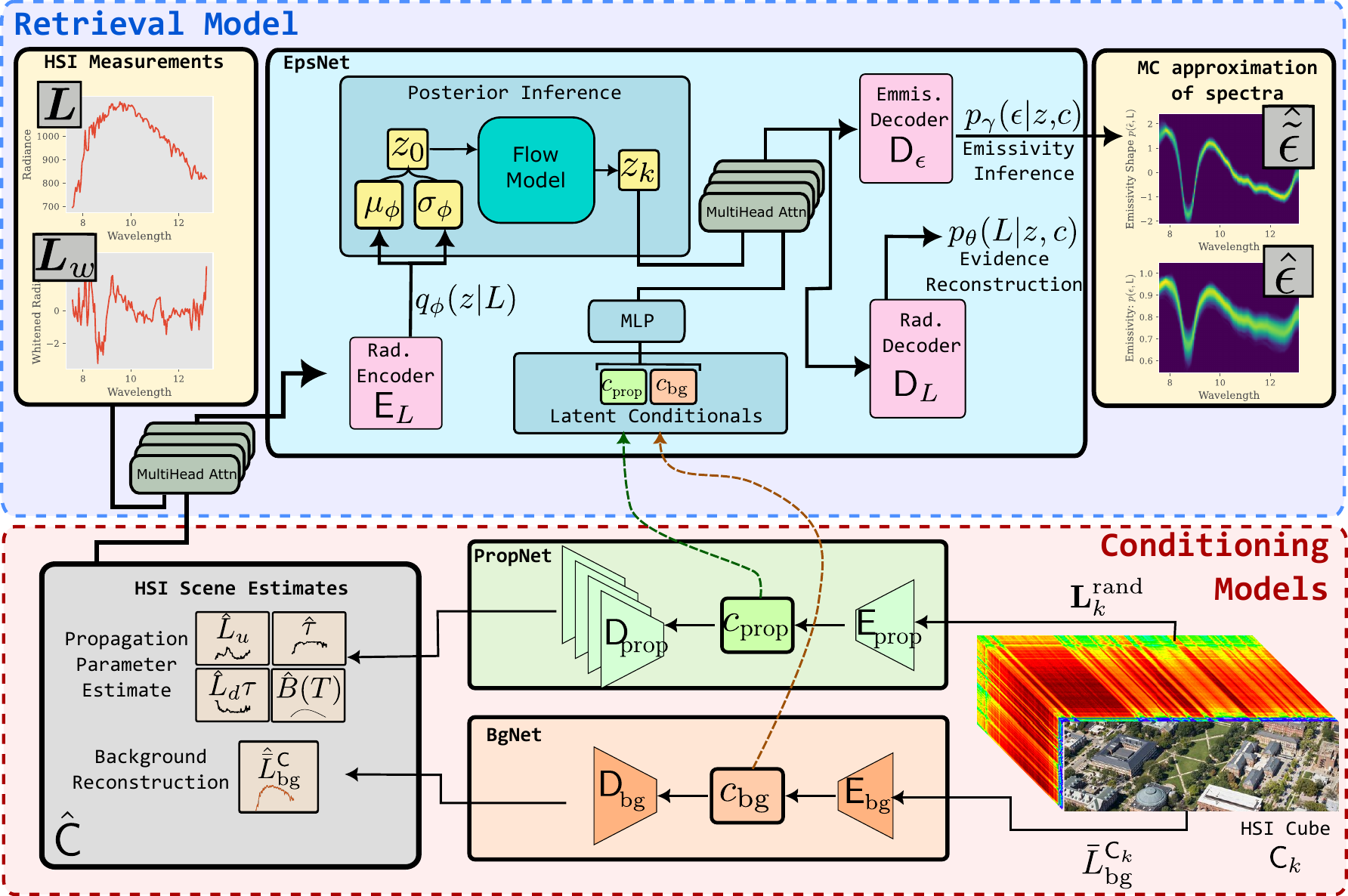}
	\caption{The data-flow and ML framework. The auxiliary networks \texttt{PropNet} and \texttt{BgNet} estimate the HSI scene parameters for a given cube, while \texttt{EpsNet} estimates $p(\eps|\rad)$ for a given radiance sample from $\cube$. Cross-attention conditioning between the pixel's radiance and HSI scene estimates applied at the input and latent spaces of \texttt{EpsNet}.
	}
	\label{Fig:ML_WorkFlow}
\end{figure*}

We take a generative modeling approach with the goal of estimating $p(\epsilon|L)$ based on a Monte-Carlo (MC) estimation using conditionally generated spectra.
The proposed modeling scheme consists of three main modules which are depicted by Fig~\ref{Fig:ML_WorkFlow} and detailed herein. The three main modules consist of a conditioned latent variable model (also referred to as the generative model or generator)
and two complementary models which serve to provide cube-wide conditional information to the generator.
The latent variable model is termed \texttt{EpsNet} and is responsible for ingesting radiance measurements and inferring the sought distribution $p(\eps|\rad)$.
The complementary networks, termed \texttt{PropNet} and \texttt{BgNet}, are two encoder-decoder networks which are responsible for estimating the propagation parameters $\mathcal{A}$ and background radiance  $\Lbgmean$, respectively.

 \texttt{EpsNet} consists of five sub-modules: the encoder $\enc_{\rad}$, a latent normalizing flow model  $\flow$, a radiance decoder $\dec_{\rad}$, an emissivity shape decoder $\dec_{\eps}$, and an emissivity scale estimator $s_{\dec_\eps}$. 
The encoder acts on the input set $\{\radset,\hat\atm,\hat\rad_{\rm bg}\}$ and produces the latent vector by inferring $\mu_\phi$ and $\sigma_\phi$ and sampling $\z\sim\mathcal{N}(\mu_\phi,\sigma_\phi)\in\mathbb{R}^{\ldim\times1}$, with $\ldim$ being the latent dimension.
The module $\enc_{\rad}$ serves to estimate the parameters of $q_\phi(\z|\radset,\hat\cube)$ which is denoted  $q^0_\phi(\z|\radset,\hat\cube)$ when sampled prior to the flow model (with corresponding latent codes $z$ or $z_0$), and $q^K_\phi(\z|\radset,\hat\cube)$ after flow operations have been applied (with corresponding latent codes $z_K$). 
Likewise, $\dec_{\rad}$ and $\dec_{\eps}$ are parameterized with $\theta$ and $\gamma$, which learn the parameterizations of $p_\theta(\rad|\z,c)$ and $p_\gamma(\eps|\z,c)$.

In what follows,   $x_{\enc_\rad}$, $x_{\dec_\rad}$, and $x_{\dec_\eps}$  denotes data flowing through the $\iter$-th layer of $\enc_\rad$, $\dec_\rad$, and $\dec_\eps$, respectively. 
The encoding blocks compress the dimension of the input from a $x_{\enc_\rad}^1\in\mathbb{R}^{128\times1}$ to $x_{\enc_\rad}^4 = \{\mu_\phi,\sigma_\phi\}\in\mathbb{R}^{\ldim\times2}$, whereas the decoding modules $\dec_{\eps}$  apply successive lifting blocks such that $x_{\dec_\eps}^1=z_k\in\mathbb{R}^{\ldim\times1}$ to $x_{\dec_\eps}^{4}=\{\hat{\tilde\eps},  \sigma_{\tilde\eps}^\lambda\}\in\mathbb{R}^{128\times2}$. The same operations are performed by $\dec_{\rad}$ which has corresponding data flows $x_{\dec_\rad}^{1}\in\mathbb{R}^{d_z\times1}$ to $x_{\dec_\rad}^{4}=\{\hat\rad, \sigma_{\rad}^\lambda\}\in\mathbb{R}^{128\times2}$.
The outputs of $\dec_\rad$ and $\dec_\eps$ are both the predicted mean and estimated variance of the radiance and emissivity, respectively, and are thus twice the dimension of the spectral length. 
The mean represents the expected value, e.g., the predicted emissivity.
The purpose of the predicted variance is to provide regularization to the error enforcement on predicted spectral shapes in terms of negative log-likelihood as opposed to direct mean-squared error, and thus encourage the model to adopt higher-variance in wavelength regions of greater uncertainty.
Between $\enc_\rad$ and $\dec_\eps$, the flow model $\flow$  acts on the sampled latent vector $\z$ to produce a transformed distribution with associated sample $\z_K$ (as detailed in section~\ref{subsec:flow}).
Altogether, the network's conditional generation of $\eps$ given $\rad$ (denoted $\eps|\rad$) is a composition of the functions,
\begin{equation}
	\hat\eps|\rad = 
	\hat\sigma_\eps\underbrace{\dec_\eps( \flow(\mathcal{R}(\enc_\rad(\radset,\hat\cube))),c )}_{\hat{\tilde\eps}|\rad } + \hat{\bar\eps},
\end{equation}
where $\hat{\tilde\eps}|\rad $ is the predicted normalized emissivity shape and $\mathcal{R}$ is the reparameterization of $\enc_\rad$'s output, e.g., $\mathcal{R}(\enc_\rad(\radset,\hat\cube))$ is equivalent to drawing samples $z \sim q_\phi(z|\radset,\hat\cube)$. 
Finally, to recover the sought empirical distribution, the conditional posterior $q_{\phi}(\z|\radset,\hat\cube)$ is sampled $N$ times and decoded to build the set of inferred emissivity measures, denoted $[\hat\eps]_N = \{\hat\eps_n|L\}_{n=1}^{N}$, leading to the empirical distribution function in normalized and scaled emissivity space,
\begin{equation}
	\begin{cases} 
		\hat{p}(\tilde\eps|L) =&   
		\frac{1}{N}\sum_{n=1}^{N}\delta(\tilde\eps-\hat{\tilde\eps}_n(z_n))\\
		\hat{p}(\eps|L) =&  
		\frac{1}{N}\sum_{n=1}^{N}\delta(\eps-\hat\eps_n(z_n))
	\end{cases},  \ \ \ 	\ z_n\sim q_\phi(z|\radset,\hat\cube)
	\label{EQ:MC_samp}
\end{equation}
where $\delta(\square)$ is a Dirac measure. We introduce the explicit dependency on $z_n$ in Eq~\eqref{EQ:MC_samp} to emphasize the relationship between the variational posterior and the inferred distributions of  $\hat{p}(\eps|L)$ and $\hat{p}(\tilde\eps|L)$.

\subsection{Posterior Estimation}
\label{subsec:flow}
To encourage expressivity in the latent distribution while still obeying the KL regularization on $q_\phi^0(z|L)$, we employ a latent normalizing flow model (NFM). 
The NFM applies successive invertible mappings to samples drawn from the base distribution $z_0\sim q_\phi^0(z|L)$ and transform them to samples of an arbitrarily complex distribution though the composition of mappings, $f_k$.
This transformation of the posterior allows the model to more accurately capture the true posterior distribution, which is certain to be beyond the complexity of the Gaussian prior $p(z)$, and thus encourages more flexible and accurate generative inference.

After the $K$-th mapping of the normalizing flow, the latent code is
\begin{equation}
	z_K = f_K\circ f_{K-1}\circ\cdots\circ f_1(z_0),
\end{equation}
with corresponding density
\begin{equation}
	q_{\phi}^{K}(\z|\rad) = q_{\phi}^{0}(\z_0|\rad) \prod_{k=1}^{K}\left|\det \bm{J}_{f_k}(z_{k-1})\right|^{-1},
\end{equation}
where $\bm{J}_{f_k} = \nabla_{\z_{k-1}} {f}_k$ is the Jacobian of the $k$-th mapping.
The learnable parameters of $\flow$ are denoted as $\kappa$ which are updated during training jointly with the other \epsnet \ modules.
We employ the real-NVP flow model in this work~\cite{Dinh2016}. Details regarding its implementation and effect on Eq~\eqref{EQ:ELBO} are given in~\ref{Apx:Flow}.

\subsection{Model Architecture}
\label{sec:architecture}

\texttt{PropNet} operates on sets of randomly sampled radiance measures of a given HSI cube, denoted as $\textbf{L}^{\text{rand}}_k \in\mathbb{R}^{200\times128}$, where the notation $\textbf{L}^{\text{rand}}_k \in  \cube_k\in\cubeset$ denotes the $k$-th hyperspectral cube, while \bgnet \ takes in the mean background $\Lbgmeank$. The \texttt{PropNet} and \bgnet \ modules encode their inputs using a deep set encoder and a standard (vanilla) encoder, respectively, so that the $k$-th cube's latent embedding of the propagation parameters and background  are,
\begin{equation} 
	\begin{aligned}
		\begin{cases}
			c_{\text{prop}}^k =  m_{\text{prop}}   \left(\frac{1}{J}\sum_{j=1}^{J}\enc_{\text{prop}}(L_j)  \right), \  L_j\sim \textbf{L}^{\text{rand}}_k  \\
			c_{\text{bg}}^k = \enc_{\text{bg}}(\Lbgmeank) 
			\end{cases}.
	\end{aligned}
\end{equation}
Here, $\enc_{\text{prop}}$ and $\enc_{\text{bg}}$ are 5-layer multi-layer perceptron (MLP) encoders which are tasked with learning the compact representation of the
propagation  and background components of a given cube $\cube$, respectively, whereas
$m_{\text{prop}}$ is a five-layer MLP that transforms the aggregated output of the deep set encoder into a more informative latent code for subsequent decoding.
The length of latent vectors $c_{\text{prop}}$ and $c_{\text{bg}}$ is denoted as $\atmdim$ and $\bgdim$, respectively, for a combined conditional latent dimension of $\cdim = \atmdim+\bgdim$, leading the overall latent representation of the HSI cube $c = \{c_{\text{prop}}\concat c_{\text{bg}}\}\in\mathbb{R}^{\cdim}$, where $\concat$ denotes feature-wise concatenation
The networks subsequently decode their respective latent representations to estimate  the unknown parameters of $\atm$ and the cube's background radiance, respectively, forming the data flows of the $k$-th cube as,
\begin{equation}
	\begin{aligned}
		f_{\text{prop}}\ :\ \	& \textbf{L}^{\text{rand}}_k \xrightarrow{\enc_{\text{prop}}} c_{\text{prop}}^k \xrightarrow{\dec_{\text{prop}}} \hat{\atm}_k\\
		f_{\text{bg}}\ :\ \	& \Lbgmeank \xrightarrow{\enc_{\text{bg}}} c_{\text{bg }}^k  \xrightarrow{\dec_{\text{bg }}} {\Lbgmeankhat}.
	\end{aligned}
\end{equation}
$\dec_{\text{prop}}$ consists of a set of 4 MLP decoders which map the propagation latent vector to the four unknown propagation parameters of $\Lt$, while $\dec_{\text{bg}}$ is a single MLP decoder which maps the background latent vector back to the cube mean. Altogether, these decoding operations account for the relevant parameters of a given HSI scene, and hence we use the notation $\hat\cube=\{\hat{\atm},\hat{L}_{\text{bg}}\}$ to denote the HSI scene estimate.
The parameterization of each auxiliary network is summarized by Table~\ref{Tab1}.

\begin{figure}\centering
	\fullfig{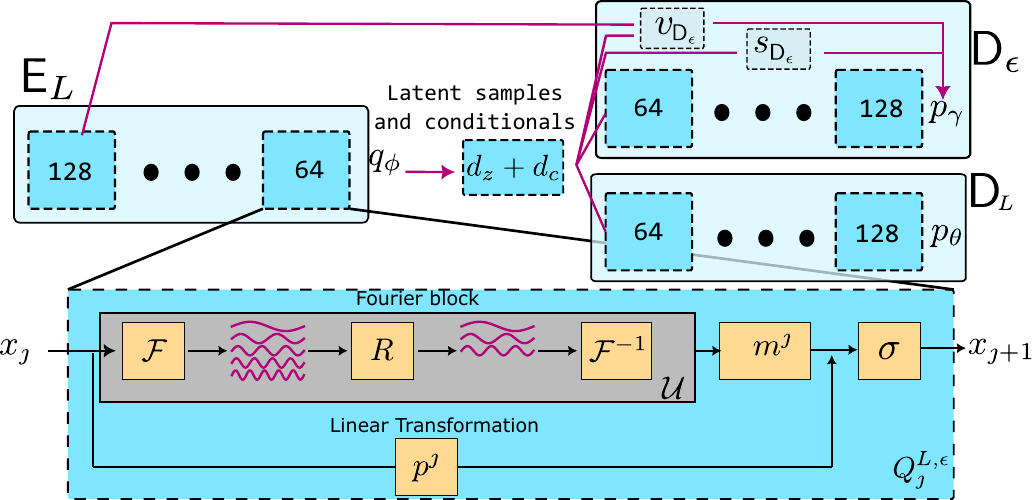}
	\caption{The encoding and decoding blocks for radiance encoding and emissivity inference. The $\iter$-th encoding and decoding block consists of FNO-MLP  blocks composed of a Fourier convolution operation $\mathcal{U}$, residual MLP network $m^\iter$, and a projection layer $p^\iter$, and output activation $\sigma$.}
	\label{Fig:FNO}
\end{figure}

The \texttt{EpsNet} encoder and decoders are comprised of four blocks, denoted in general as $Q_\jmath^{\square}$ for the $\jmath$-th block of network $\square$, with each following a Fourier-Neural Operator (FNO) - MLP architecture depicted in Fig~\ref{Fig:FNO}. We denote the input tensor to the $\iter$-th block of \texttt{EpsNet} as $x^\iter$. Accordingly, the block-wise operation of encoding and decoding is expressed as the composition
\begin{equation}
	\begin{aligned}
		\enc_{\rad} &= Q^{\enc_\rad}_1\circ\cdots\circ Q_N^{\enc_\rad}\\
		\dec_{\rad} &= Q^{\dec_\rad}_1\circ\cdots\circ Q_N^{\dec_\rad}\\
		\dec_{\eps} &= Q^{\dec_\eps}_1\circ\cdots\circ Q_{N}^{\dec_\eps}\\
		Q_{\iter}^{\square}(x^\iter) 
		&= p^\iter(x^\iter ) +  m^\iter(\mathcal{U}(x^{\iter-1})).
	\end{aligned}
\end{equation}
Here, $p^\iter$ is a linear projection layer which aligns to the output dimension of $x^{\iter+1}$ enabling an addition residual operation, $m^\iter$ is a four-layer residual-net (ResNet) MLP comprised of fully connected layers, and $\mathcal{U}(x^\iter)$ is a spectral convolution layer,
\begin{equation}
	\mathcal{U}(x) = \int \varsigma (x,\xi)v(\xi){\rm d}\xi = \mathcal{F}^{-1}\left(R_\chi \mathcal{F}(v)\right)(x),
\end{equation}
where $\mathcal{F}\{\square\}$ is the Fourier Transform of $\square$, and $\varsigma$ is a representative kernel function which is learned in the frequency domain by the  linear transformation $R_\chi$ operating on a reduced basis of Fourier modes.
The primary motivation for employing Fourier layers is that the spectral decomposition captures smooth global representations of the data in an infinite-dimensional function space, enabling the model to better learn compact latent representations for tasks involving functional transformations~\cite{Li2020,Kovachki2023}.  

The primary decoding blocks of $\dec_\eps$ learn the normalized emissivity, $\tilde\eps$, such that each spectral sample is z-standardized per its own statistic. 
This is done to avoid bias towards spectra with high amplitude features, as low-variance spectra (which are common in the available library) are difficult to learn from data-driven error metrics alone.
For this reason, the normalized emissivity  \textit{shape} is predicted separate from its mean and standard deviation, $\hat{\bar\eps}$ and $\hat\sigma_\eps$. 
In this sense, the errors in misrepresenting spectral features are equally applied to each sample irrespective of a material's relative variations within the LWIR bands.
An additional module within $\dec_\eps$ which is termed \texttt{ScaleNet} is tasked with inferring the spectral mean and standard deviation of each sample based on samples from the posterior,
\begin{equation}
	\{\hat{\bar\eps},\hat\sigma_\eps \} = s_{\dec_\eps}(\{\z_K\concat c\}), \ \ \z_K\sim q_\phi^K(\z|\radset,\hat\cube).
\end{equation} 
The architecture of \texttt{ScaleNet} is a 5-layer MLP with hyperbolic tangent activation that maps $\{\z_K\concat c\}\in\mathbb{R}^{\ldim+\cdim}$ to $\{\hat{\bar\eps}\concat\hat\sigma_\eps \}\in\mathbb{R}^{2}$, with all hidden layers possessing 128 dimensions. 
To estimate variance in the LWIR bands, e.g., learn where to allow for greater uncertainty in the inverse mapping by assigning wavelength-dependent leniency in misrepresenting spectral features, an additional network termed \texttt{VarianceNet} $v_{\dec_\eps}$ operates on the input sequence to $\enc_\rad$ and conditioned latent sequence, 
\begin{equation}
	\sigma_{\tilde\eps}^{\lambda} = v_{\dec_\eps}\left(\{	x^{1}_{\enc_\rad}\concat\z_K\concat c
	\}\right), \ \ \z_K\sim q_\phi^K(\z|\radset,\hat\cube).
\end{equation}
The architecture of \texttt{VarianceNet} is a 5-layer MLP with each hidden layer possessing 128 dimensions.
The parameters of \texttt{ScaleNet} and \texttt{VarianceNet} are denoted as $\varpi$.
For each MLP sub-module in $\enc_\rad$, $\dec_\eps$, and $\dec_\rad$, the hidden dimension matches that of $x^\iter$, while the last layer and projection layer compresses or expands the data to the output dimension of $x^{\iter+1}$. 
We utilize self-gated activation in each decoding network (commonly known as Swish activation), and Sigmoid activation in the encoding network. 
The free network hyperparameters were taken as the latent dimension $\ldim=64$, the number of blocks $n_b=4$, and the number of MLP layers per block $n_l=4$; these parameters were determined via Bayesian optimization using a Matern 5/2 kernel and the maximum expected improvement acquisition function~\cite{brochu2010tutorial}.

\begin{table}[t!]
	\centering
	\caption{Parameterization of Auxiliary Networks}
	\label{Tab1}
	\begin{tabular}{lll}\hline
		Parameter 					& Description & Value\\\hline
		$\ldim$ 						& \texttt{EpsNet} Latent dimension, $\dim(z)$ & 64	\\ 
		$\atmdim$  					&  \texttt{PropNet} Latent dimension & 12 			\\ 
		$\bgdim$ 					&  \texttt{BgNet} Latent dimension 	& 12				\\ 
		$\sigma_{\text{prop}}$ 	& \texttt{PropNet} Activation & Tanh		\\
		$\sigma_{\text{bg}}$ 	& \texttt{BgNet} Activation & Tanh		\\
		$d_{h_\text{prop}}$ 		& \texttt{PropNet} Hidden Dim 	& 128		\\
		$d_{h_\text{bg}}$  		& \texttt{BgNet} Hidden Dim 		& 	128		\\
		\hline
	\end{tabular}
\end{table}

\subsection{Physics-Based Conditioning Mechanisms}
\label{subsec:conditioning}

The conditional networks serve to provide contextual information to the \texttt{EpsNet} based on the propagation model.
The goal is to mitigate the ill-posedness and over-parameterization of Eq~\eqref{eq:fp}'s inverse, 
allowing \epsnet \  to contextualize a single-pixel measurement based approximations of $\atm$ and $\Lbgmean$ provided by the auxiliary network.
%
The conditioning to the input layer of \texttt{EpsNet} is applied via a cross-attention (CA) module between the  estimated parameters and the observed radiance, while the decoder inputs are conditioned by the CA  between the latent codes sample from the posterior distribution and the latent representation of the cube parameters,
\begin{equation}
	\begin{aligned}
		\encodervae^1  &= \text{CA}(\radset, \{\hat{\atm}\concatc  \hat{\rad}_{\text{bg} } \})  =\text{CA}(\radset, \hat\cube ) ,\\ 
		\epsnetvae^1,\decodervae^1 	&= \text{CA}(z_K, \{c_{\text{prop}}\concatc  c_{\text{bg} } \}))  =\text{CA}(z_K, c ) ,\\ 
	\end{aligned}
\end{equation}
where $\concatc$ is channel-wise concatenation.
In this scheme, the sensor measurement set $\radset$ and its latent representation serve as the query set that possesses the primary information pertinent to emissivity retrieval, while the estimates of the HSI scene and their respective latent representations serve as the keys which possess contextual information about $\radset$.
This allows \texttt{EpsNet} to produce conditional posterior estimates that account for the unknown variables in Eq~\eqref{eq:fp}, as well as conditioned emissivity inference. 
In the input attention case, a final mean-aggregation is taken to produce a single vector to \texttt{EpsNet} for each observed pixel.

\subsection{In-the-loop Augmentation}
\label{subsec:epsmod}
To encourage the model to learn a more general inverse mapping, as opposed to recalling a set of $M$ predefined spectral signatures available in $\epsset$, the training and testing emissivity need to be sufficiently diversified. 
To ensure this, the components of $\epsset$ were modified at each training epoch so that no two target spectra are ever the same during training. 
Because on-the-fly emissivity generation is currently infeasible with laboratory measurements and intractable via first-principals modeling, a synthetic emissivity manipulation scheme was adopted. 
Moreover, since the overarching goal of the model is to produce the map $f:\radfspace\to\epsfspace$, modifications of $\eps$ are not strictly required to be constrained to real-world spectra. 
Nevertheless, the applied perturbations should leave the statistical distribution of the emissivity mean, smoothness, and variance relatively unaffected so that the perturbed spectra emulate realistic spectral shapes and thereby encourage the model to bias towards realistic spectral profiles without memorizing specific spectral of the available training data.

\begin{figure}\centering
	\fullfig{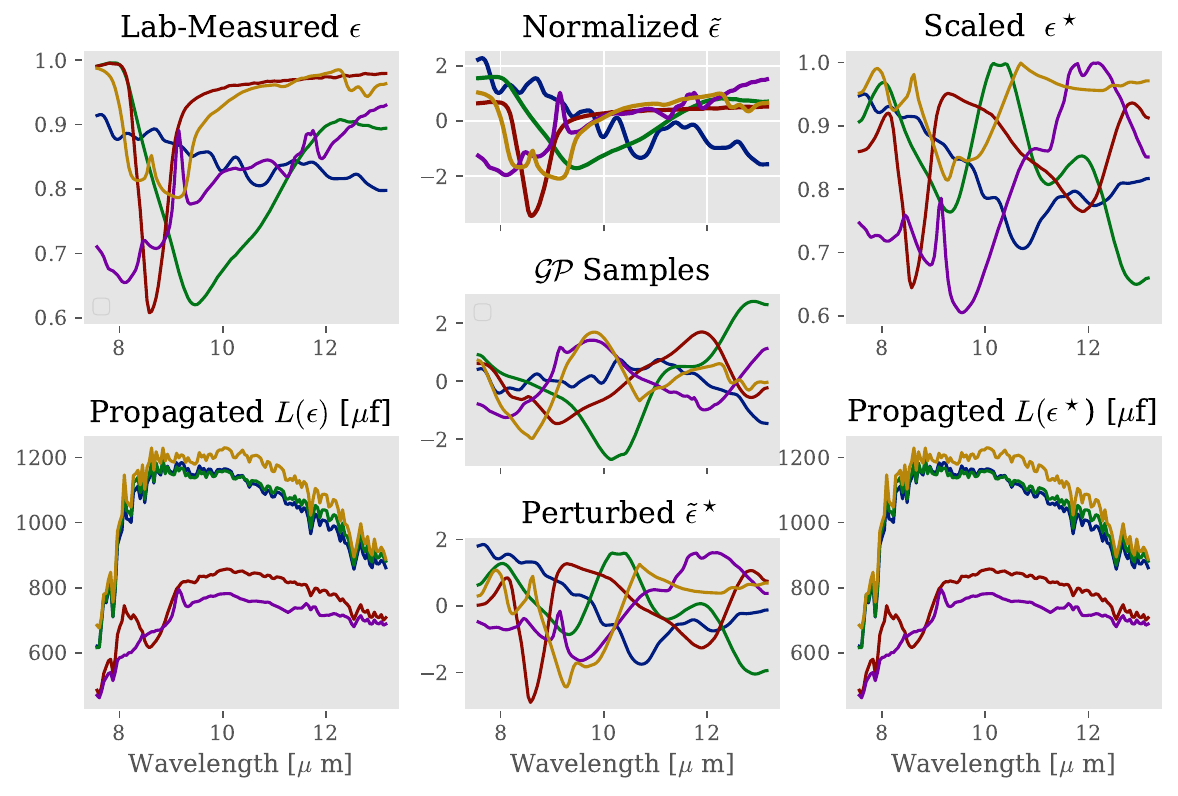}
	\caption{In-the-loop Data Augmentation. Left: Laboratory measured spectra and their respective radiance measures as taken directly from training scene in $\cubeset$. Middle: The normalization of the spectral data, samples from a Gaussian process, and normalized perturbed spectra. Right: The scaled perturbed spectra and its re-propagated radiance, serving as the \textit{novel} batch sample. }
	\label{Fig:EpsMod}
\end{figure}

\begin{figure}\centering
	\fullfig{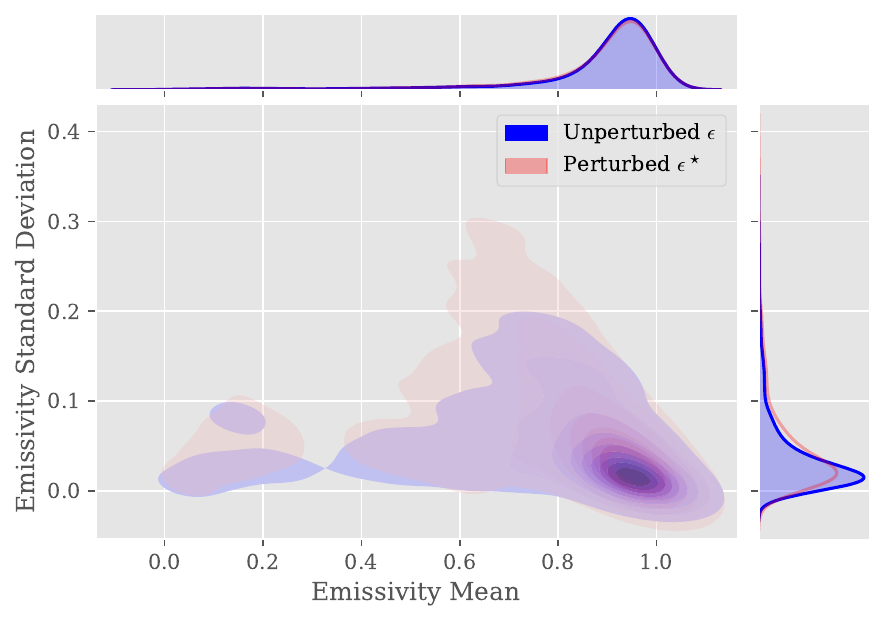}
	\caption{Comparative statistics of the mean and standard deviations between the unperturbed (lab-measured) spectra, $\eps$, and the modulated (perturbed) spectra $\eps^\star$. Results are given for a batch sample of size 1024.}
	\label{Fig:EpsModStat}
\end{figure}

The emissivity perturbation scheme is described here and depicted by Fig~\ref{Fig:EpsMod}.
First, the emissivity measures are each $z$-standardized with respect to their own statistics so as to ``normalize'' the relative effect of subsequent perturbations, $\tilde\eps = (\eps-\bar\eps)/\sigma_\eps$.
To ensure sufficient smoothness and spatial structure, the perturbations are applied in the form of a Gaussian Process ($\mathcal{GP}$) that is parameterized by an expected value of 0 and covariance kernel $k_{xx}$ which we take as a Matern 5/2 kernel,
\begin{equation}
	\tilde{\eps}^\star_k = \bar{\eps}_k + \mathcal{GP}(0,k_{xx}).
\end{equation}
Once modified, the emissivity is re-scaled per $\eps^\star =\text{softclamp} ( \tilde{\eps}^\star\sigma_\eps + \bar\eps)$
where the softclamp operation ensures the modulated emissivity falls between 0 and 1. The clamped spectra are re-normalized as well to provide the set $\{\eps^\star,\tilde\eps^\star\}$.  The quantity $\eps^\star$ is then re-propagated per Eq~\eqref{eq:fp} using the same atmosphere, background, and whitening parameters for the given radiance observation so as to return the propagation of the novel sample into the same scene, $L^\star = f_{\rm p}(\eps^\star;\atm,\Lbg)$.

Fig~\ref{Fig:EpsMod} depicts an example of the emissivity modulation procedure for a mini batch of five pairs of $(\eps,\rad)$ (shown in the left-column). The $\mathcal{GP}$ is sampled and added to the normalized samples in the center column producing $\hat\eps^\star$, before being re-scaled per the batch's original statistics and re-propagated via Eq~\eqref{eq:fp} as shown in the right column.
Fig~\ref{Fig:EpsModStat} depicts the preservation of the batch statistics for a batch of 1024 samples, showing the distributions of the mean and standard deviation  of  emissivity samples before and after modulations. 
The distributions of $\eps$ and $\eps^\star$ are nearly identical. Moreover, a very high concentration of spectra have nearly unit mean and nearly zero variance, indicating a prominence relatively flat (low-variance) spectral bands, with few high-variance spectra included.
After normalization, however, all spectra are statistically equivalent in terms of shape learning via $\dec_\eps$.

\subsection{Training Scheme}


The model training consists of two separate stages: conditional model training and inference model training. The former is responsible for training \atmnet\ and \bgnet\ in order to first develop models capable of providing the necessary conditioning to \epsnet; it is performed offline prior to training \epsnet. The latter is responsible for training the learnable parameters of \epsnet.

\subsubsection{Conditional Network Training}
To train \atmnet, 2000 HSI cubes were collected into a dataset with corresponding propagation parameters to form the training dataset ($\atmset_\atmnet$,$\cubeset_\atmnet$). From $\cubeset_\atmnet$, 100,000 radiance sets (with a single set here denoting $\textbf{L}^{\text{rand}}_k \in\mathbb{R}^{200\times128}$) were drawn to form training and validation datasets.
A joint training objective was adopted. The first is standard MSE loss between  $\hat\atm$ and  $\atm$, while the second is a propagation objective  whereby a flat emissivity curve of $\eps=0.5$ with unit strength is used to propagate the data per Eq~\eqref{eq:fp}. 
\begin{equation}
	\begin{aligned}
	\mathcal{L}_{\text{prop}}(\varrho) = \frac{1}{K} \sum_1^k\Big[ & (1-\beta) || \mathcal{A}-\hat{\mathcal{A}}|| _2 ^2 +
	\\ 
	&\beta||  f_{\text{p}}(0.5,\mathcal{A},1)- f_{\text{p}}(0.5,\hat{\mathcal{A}},1)|| _2^2\Big]  
\end{aligned}
\end{equation}
The first loss metric provides a direct enforcement of the atmospheric reconstruction.
The latter loss provides a physically meaningful measure with respect to the end goal of emissivity retrieval, as it encourages \atmnet \ to learn a parameter set $\varrho$ and latent representation that provides the closest match to true atmospheric composition in terms of the forward propagation process. 
We note that the selected value of $\eps=0.5$ is lower than average for most materials (see Fig~\ref{Fig:EpsModStat}). However, this was selected during training to prevent biasing \atmnet\ toward de-emphasizing accurate inference of $L_d$ relative to $B(T)$ (see Eq~\eqref{eq:fp}).
Training is scheduled by the relative weight $\beta$ which scales linearly from 0-to-1 throughout training over 1000 epochs with a learning rate of $9\times10^{-4}$.

To train \bgnet, the set of 2000 training cubes and their outlier-removed means were utilized, ($\bgset_\atmnet$,$\cubeset_\atmnet$).
A standard MSE error was applied during training to learn the \bgnet\ parameter set $\upsilon$ over 1000 epochs with a learning rate of $10^{-3}$.

\subsubsection{\texttt{EpsNet} Training}

\begin{figure}[t!]\centering
	\fullfig{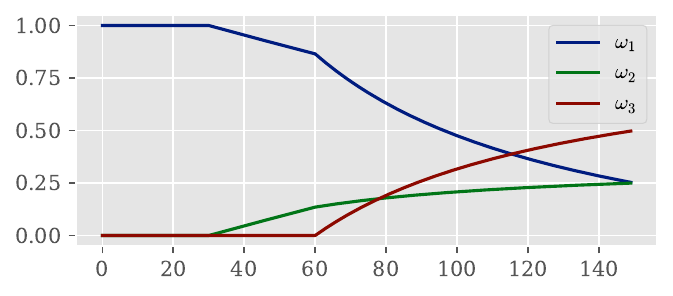}
	\caption{Relative weights over training epochs.}
	\label{Fig:TrainSchedule}
\end{figure}

The process of learning the inverse map, e.g., inferring $p(\eps|\rad)$, is guided by $\mathcal{L}_{\text{inversion}}$ which comprises six different components aimed at (i) learning the shape of the normalized spectra and enforcing a measure of smoothness (which is motivated by the continuity requirement for real emissivity curves), (ii) learning the relative scaling of the sample, e.g., it's mean and standard deviation, and (iii) enforcing a direct error metric before and after scaling the spectra per the mean and standard deviation estimates.
In the normalized emissivity space, the error computed as the negative log-likelihood of the heteroscedastic Gaussian, with the  wavelength-dependent variance and mean predicted by $\dec_\eps$.
Altogether, the composite loss function is given as,
\begin{equation}
		\mathcal{L}_{\tilde\eps}  = 
		\underbrace{\mathcal{L}_{\rm shape} +
		\mathcal{L}_{\rm smooth}
	}_{(i)}		+\underbrace{\mathcal{L}_{\rm sdev} +
	\mathcal{L}_{\rm mean}
	}_{(ii)}		+\underbrace{\mathcal{L}_{\tilde\eps} +
	\mathcal{L}_{\eps}
	}_{(iii)}.
\end{equation}
For a batch size of $K$ with spectral length $J$, the specific form of each sub-loss is expressed as,
\begin{align}
			\mathcal{L}_{\rm shape} 		=&  \frac{\omega_{\rm shape}}{K}	\sum_{k=1}^{K}	\frac{{\tilde\eps}_{k}\cdot\hat{{\tilde\eps}}_{k}}
			{\big\| {\tilde\eps}_{k} \big\| \big\|\hat{{\tilde\eps}}_{k}\big\| }\\
			\mathcal{L}_{\rm smooth} 			=& \frac{\omega_{\rm smooth}}{K(J-2)}\sum_{k=1}^K\sum_{j=0}^{J-3}(\eps_{k,j+2} - 2\eps_{k,j+1}+\eps_{k,j})^2
			\\
			\mathcal{L}_{\rm sdev} 			 	=&\frac{\omega_{\rm sdev}}{K}	\sum_{k=1}^{K}\left|\left|\hat\sigma_{\eps_k} - \sigma_{\eps_k}\right|\right|_2^2	
		\end{align}
		\begin{align}
			\mathcal{L}_{\rm mean} 		 	 	=& \frac{\omega_{\rm mean}}{K}	\sum_{k=1}^{K}\left|\left|\hat\mu_{\eps_k}	- 	\bar{\eps_k}\right|\right|_2^2 	
			\\
		\mathcal{L}_{\bar{\epsilon}} =& \frac{\omega_{\bar{\epsilon}}}{K} \sum_{k=1}^{K} \sum_{j=1}^{J} \left[ \log  \left( \sigma_{\tilde{\epsilon},j}^{\lambda} \right)^2_k  + \frac{ \left( \hat{\tilde\epsilon}_{k,j} - \tilde{\epsilon}_{k,j} \right)^2 }{ \left( \sigma_{\tilde{\epsilon},j}^{\lambda} \right)^2_k } \right]
			\\
			\mathcal{L}_{\eps} 				 	=& 	\frac{\omega_{\eps}}{K} \sum_{k=1}^{K}		\left|\left|\hat{\eps}_{k} - \eps_{k}\right|\right|_2^2,
\end{align}
where $\omega_{\square}$ is a relative scaling term for $\mathcal{L}_{\square}$.
Moreover, by propagating the scaled emissivity predictions back into radiance space per Eq~\eqref{eq:fp} (with the ground truth measures of $\atm$ and $\Lbgmean$ begin assumed available during training), we introduce a propagation loss $\mathcal{L}_{\rm propagation}$ which provides a measure of how consistent a predicted spectra  is with the observed radiance. For a batch of $J$ unique cubes and with $K_j$ radiance examples drawn from the $j$-th cube, the loss is calculated as,
\begin{equation}
	\mathcal{L}_{\rm propagation} =  \frac{1}{KJ}\sum_{j=1}^{J}\sum_{k=1}^{K_j} \left\|  f_{\rm p}(\hat\epsilon_{k,j}; \mathcal{A}_j,\alpha_k) - L_{k,j}\right\| _2.
\end{equation}
Lastly, we enforce the regularization loss, e.g., the KL-divergence objective of the ELBO loss, which is modified to account for the effect of the normalizing flow on the estimated posterior,
\begin{equation}
	\begin{aligned}
		\mathcal{L}_{\text{regularization}} =& \kld\left(q_\phi(\z|\rad)|| p(z)\right)\\&-	
		\sum_k\mathbb{E}_{z_0\sim q_\phi^0(z_0|\rad)}	\log \left|	\det J_k(z_k)	\right|.
	\end{aligned}
\end{equation}
 Altogether, the joint training objective of \epsnet\ is expressed as,
\begin{equation}
	\mathcal{L}(\bm{\vartheta},\bm{\omega}) = 
	\omega_1\mathcal{L}_{\text{inversion}} 			+ 
	\omega_2\mathcal{L}_{\text{propagation}} 			+ 
	\omega_3\mathcal{L}_{\text{regularization}}
	\label{EQ:TrainingFunction}
\end{equation}
where $\bm{\vartheta} = \{\theta,\phi,\gamma, \kappa,\varpi \}$ denotes the set of parameters for the various sub-modules of \epsnet\  which are jointly updated per the common objective Eq~\eqref{EQ:TrainingFunction}.
During training, we enforce a weight schedule to $\omega_1$, $\omega_2$, and $\omega_3$ which is depicted by Fig~\ref{Fig:TrainSchedule}. The schedule  encourages the model to first emphasize the inversion map in a deterministic fashion (e.g., with regularization and propagation losses ignored). Next, the weight of the propagation loss is slowly introduced, so that the model is encouraged to tailor the inverse predictions to be as physically consistent as possible with the observed radiance. Finally, the regularization is activated to enforce finite latent variance leading to the sought probabilistic mapping.

The \epsnet\ modules were jointly optimized using a learning rate of $10^{-3}$ for 150 epochs with an exponentiation weight decay coefficient of 0.99 and a weight-regularization constant of $5\times10^{-5}$.
We utilized 2,500,000 radiance samples during training by injecting spectral signatures from a lab-measured library into 2,000 HSI scenes, leading to  approximately 1,200 samples per cube. 
Further details of the dataset design are given in section~\ref{subsec:dataset}.
The modulation routine described in section~\ref{subsec:epsmod} was applied at every epoch with $\mathcal{GP}$ samples generated with unit variance.

\section{Numerical Experiments and Results}
\label{SEC:Results}



To evaluate the proposed modeling scheme, we consider series of numerical experiments to investigate the model's effectiveness for probabilistic emissivity retrieval.
Moreover, we compare the proposed model to an equivalent one that is not privy to physics-based conditioning, e.g., without access to the HSI scene estimates and cross-attention scheme described in section~\ref{subsec:conditioning}. 
In what follows, we describe the datasets used for training and testing (section~\ref{subsec:dataset}), overview the learning performances of both models (section~\ref{subsec:learning}), evaluate their accuracy in a variance-aware fashion  (section~\ref{subsec:stat_eval}), and demonstrate the practical use of the model as a probabilistic material ID tool  (section~\ref{subsec:material_id}).

\begin{figure}\centering
	\fullfig{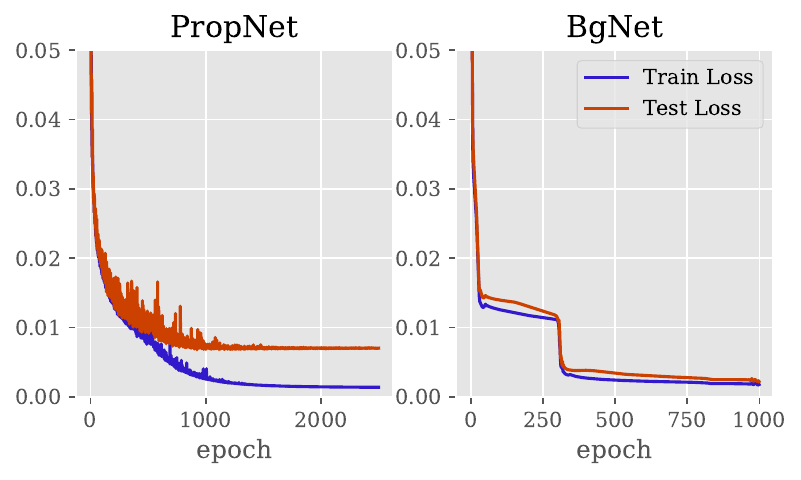}
	\caption{The loss curves of (left) \atmnet\ and (right) \epsnet\ training.
	}
	\label{Fig:MetaLoss}  
\end{figure}

\begin{figure*}[t!]\centering
	\fullfigdouble{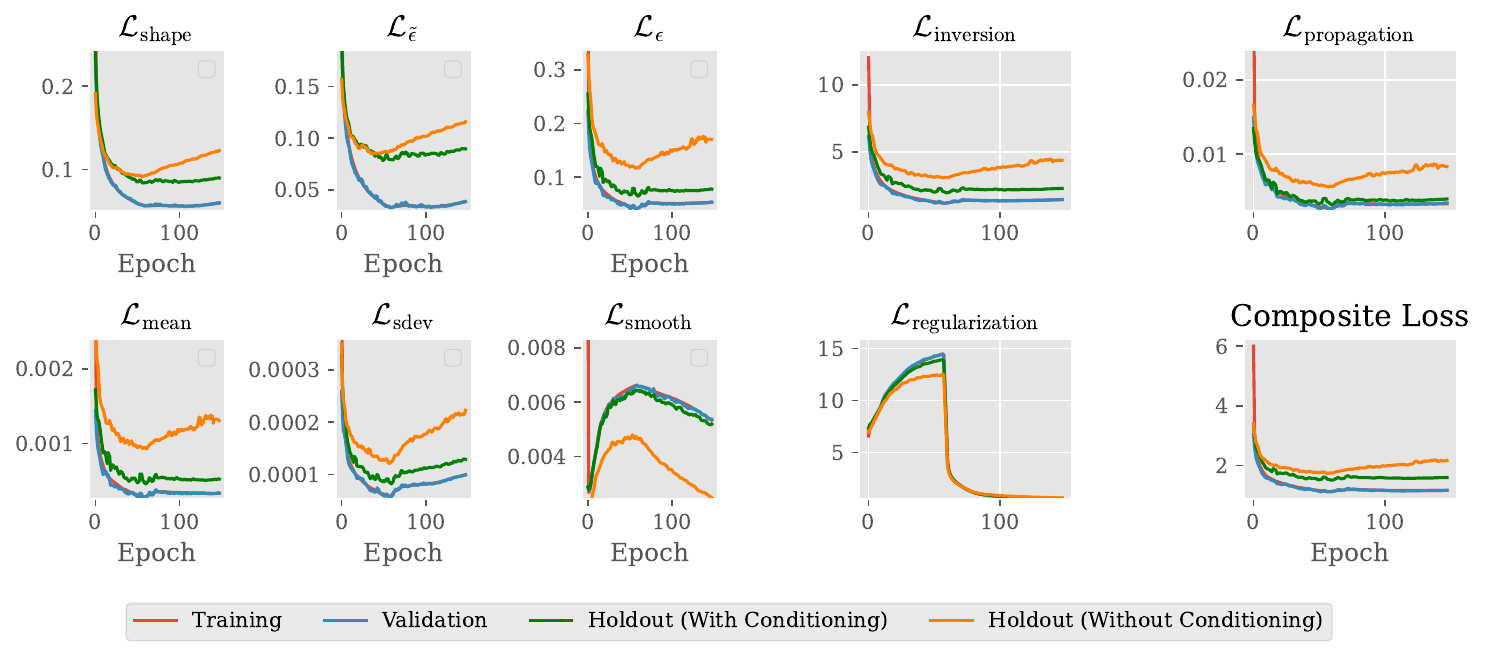}%
	\caption{Loss curves for the training objective defined by Eq~\eqref{EQ:TrainingFunction} depicting training, validation, and testing data, as well as the test loss of the model not privy to conditional data streams.
		The left three columns depict the itemized components of $\mathcal{L}_{\text{inversion}}$, whereas the right two columns depict the three over-arching objectives and their normalized combination.}
	\label{Fig:Loss}
\end{figure*} 
\subsection{Dataset Design}
\label{subsec:dataset}
Two datasets of hyperspectral data were constructed to train, validate, and test the model over. 
The measurements of each scene were collected with a Mako airborne sensor as described in~\cite{Buckland2017}, with measurement dates spanning between 2018 and 2024.
The sensor resolution of each measurement is fixed at 128 wavelength bands between 7.56 and 13.16 $\mu$m.
Having access to a library of available HSI cube measurements spanning a diverse set of geographical locations, dates, and times, we partitioned a training library of $K_{\text{train}}=2,000$ HSI cubes along with their estimated atmospheric variables $\atm$ and background $\Lbgmean$  per conventional physics-based atmospheric compensation.
The training library was constructed to only include data collected on even-valued days.
To ensure that the training and validation data did not overlap with hold-out (test) data, a separate library of HSI data was constructed consisting of  $K_{\text{test}}=500$ cubes with measurements made on odd-valued dates.
In each library, a proprietary set of  $M=$10,095 laboratory measured LWIR emissivity curves for solid materials was injected into the scene, with the parameters $\atm$ propagating the target radiance ($\Lt$ of Eq~\eqref{eq:fp}) for the pixel, and with the pixel's measured background radiance accounting for $\Lbg$.

We numerically explored the performance of the trained model by generating MC-approximations of $p(\eps|\rad)$ per Eq~\eqref{EQ:MC_samp}. 
For a given radiance observation, an inference set size of 1,280 samples (10$\times$ the dimensionality) was generated to construct $\hat{p}(\eps|L)$, which is  sufficient  for constructing well-approximated covariance matrices of the empirical distribution.
The numerical evaluation of the model was computed for test data that was not introduced to the model during training, e.g., the hold-out set.
The target strengths ranged between $\alpha=0.1$ and $\alpha=1.0$ for training and testing sets unless otherwise stated.

\subsection{Learning Performance}
\label{subsec:learning}
The training and validation losses of the auxiliary networks  (\atmnet \ and \bgnet) and primary network (\epsnet) are depicted in Figs~\ref{Fig:MetaLoss}  and~\ref{Fig:Loss}, respectively. 
Starting with \atmnet, we note that over-training begins near the 1,000 epoch mark. However, this occurs after sufficient learning of the atmospheric makeup, as is confirmed by the reconstruction examples presented in~\ref{Appendix:Atm}. Hence, despite the seemingly disparate numerical performance of Fig~\ref{Fig:MetaLoss}, \atmnet\ sufficiently captures the dominant features of propagation components, resulting in meaningful conditioning information for the modules of \epsnet.  
Over-fitting is less apparent in \bgnet\ training, and this is due to the relative simplicity of \bgnet's learning objective of auto-encoding a 128-dimensional radiance vector. In summary, the training history  indicate a meaningful embedding of $c_{\text{bg}}$ was learned for the HSI scene background.

The learning results of Fig~\ref{Fig:Loss} are the more germane for the overall modeling goal, with inversion, propagation, and regularization objectives being highlighted separately for a holistic evaluation of the objective~\eqref{EQ:TrainingFunction}.
For each plot, the training loss, validation loss (e.g., in-distribution hold-outs), testing loss (out-of-distribution hold-outs), and testing loss without physics conditioning are shown.
The left six plots depict the six itemized loss functions of   $\mathcal{L}_{\text{inversion}}$. 
During the initial 60 epochs, before regularization is activated, the loss curves for  $\mathcal{L}_{\text{propagation}}$ and  each component of $\mathcal{L}_{\text{inversion}}$ fall significantly (with the exception of smoothing loss) as the network learns a pseudo-deterministic inverse map. 
During this burn-in stage, the test loss of conditioned and unconditioned networks are comparable for shape-based metrics, while the reconstruction based metrics (e.g., $\mathcal{L}_{\text{mean}}$,  $\mathcal{L}_{\text{sdev}}$,  $\mathcal{L}_{\eps}$) show considerable deviation between the physics-conditioned and unconditioned models. 

Prior to the regularization objective being activated, the KL-divergence expectedly grows as the posterior is pushed towards zero variance. Upon its activation, a sharp drop in $\mathcal{L}_{\text{regularization}}$ is observed, which eventually settles at a sufficiently low value ($\approx 0.5$ for training and testing), indicating that the unit-variance prior is sufficiently matched and an appreciable level of uncertainty has been introduced to the mapping. 
Notably, the inversion and reconstruction losses remain relatively flat for the physics-conditioned model after the regularization enforcement, indicating that the model retains a majority of its capacity for performing the mapping even under the latent uncertainty. Since millions of examples are evaluated at each epoch, the reconstruction losses of the regularized model are indicative of the error between the target emissivity measures and the model's expectation under sufficiently large sampling.
In contrast, these same loss quantities grow for the unconditioned model upon regularization enforcement, indicating that the unconditioned model struggles to simultaneously satisfy the reconstruction and regularization objectives.

\begin{figure*}[t!]\centering
	\fullfigdouble{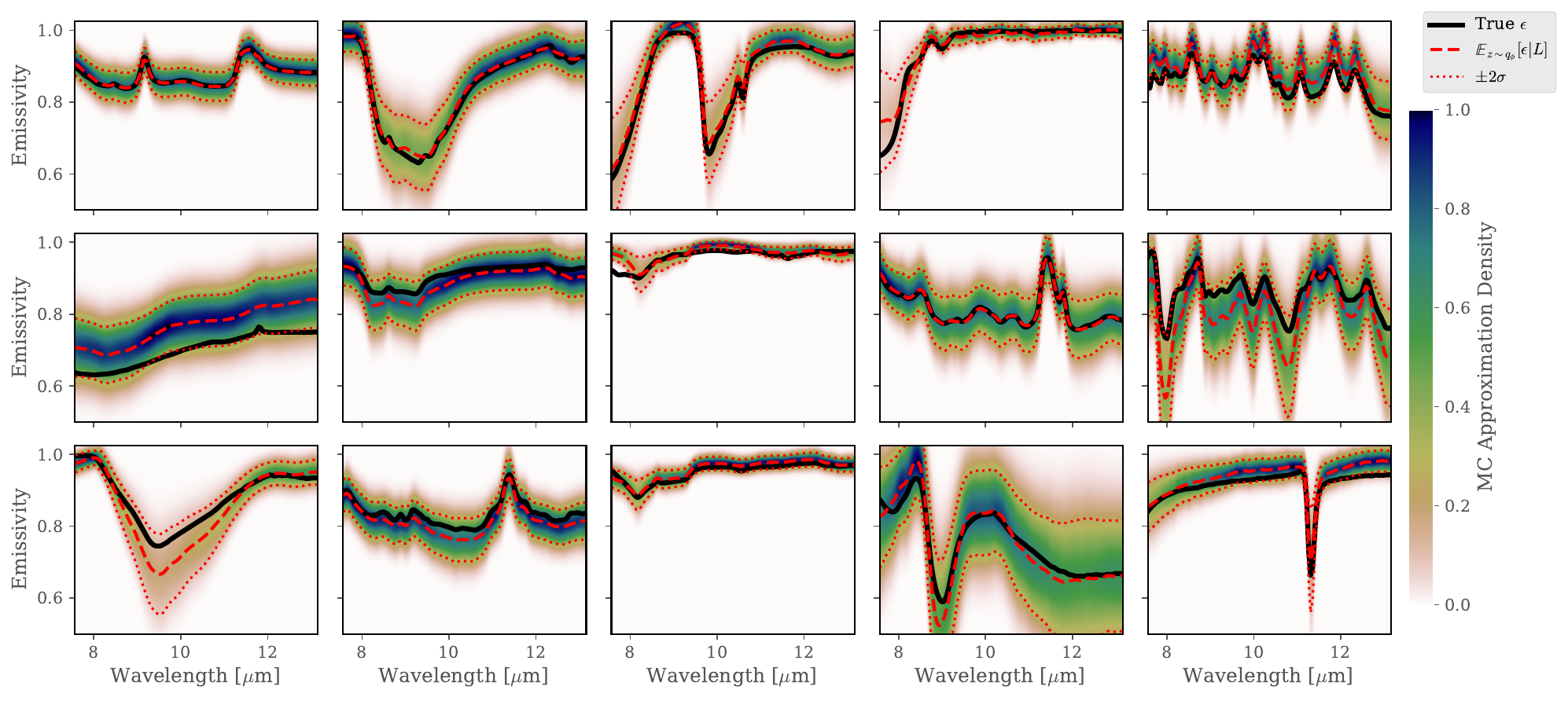}
	\caption{Visualizations of $\hat{p}(\eps|\rad)$. The color map depicts the kernel density estimates for sampled sets of emissivity of size 1280, while the thick and thin dashed-red lines depicted the empirically estimated expectation and $2\sigma$ variance threshold, respectively. The ground truth (e.g., the emissivity which was propagated into the observed radiance measure) is depicted with a thick black line.}
	\label{Fig:KDE}
\end{figure*}

From the above results, we deduce the following: (i) that 
the whitened data allows for the relatively accurate learning of spectral shape (with or without physics conditioning), (ii) that conditional information $\hat\cube$ is highly potent for contextualizing the relative magnitude of the $\Lt$ given an observed radiance measure, and (iii) that the physics contextualization is important for the model to jointly satisfy both the inversion and regularization objectives.
Briefly expanding on (ii), we note that the amplitude of $\Lt$ is a function of both the temperature $B(T)$ and relative pixel fill $\alpha$. Hence, by conditioning on the HSI scene, the model implicitly learns the relative prominence of $\Lt$ within $\rad$ by contextualizing the observation with the estimated black-body function and atmospheric parameters of each cube. 

\subsection{Statistical Evaluation}
\label{subsec:stat_eval}

We now seek to evaluate the empirical distributions generated by MC-sampling per Eq~\eqref{EQ:MC_samp}.
We begin with a qualitative evaluation of the generated empirical emissivity distributions with kernel density estimates (KDE) of MC samples.
Fig~\ref{Fig:KDE} depicts the resulting KDE of scaled emissivity, $\text{KDE}(\{\eps_n|L\}_{n=1}^{N})$, for 18 randomly selected test samples.
The ``ground truth" emissivity, e.g., the emissivity which was propagated through Eq~\eqref{eq:fp} to generate the input radiance measure, is superimposed with a thick black line. For the majority of samples shown, this line falls well-within the high-density regions of the KDE indicating a good match to the inferred distribution. The expected value of the MC approximation (denoted $\mathbb{E}_{z\sim q_\phi}[\eps|\rad]$) and the $2$-$\sigma$ variance bounds are also plotted, with good agreement being shown between the expected value of the inferred distribution and target emissivity which accounted for the input radiance measurement.
Hence, Fig~\ref{Fig:KDE} demonstrates (in a qualitative sense) that the goal of estimating a conditional distribution $\hat{p}(\eps|\rad)$ carrying finite uncertainty is achieved by the model.

\begin{figure}\centering
	\fullfig{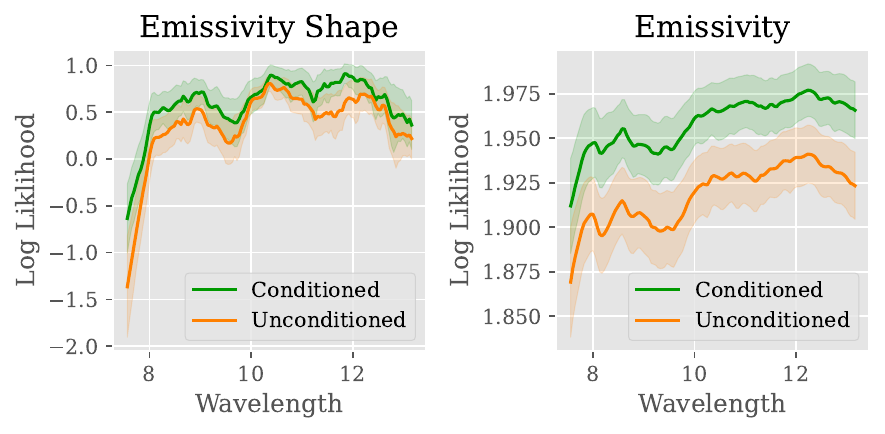}
	\caption{Wavelength-dependent likelihood estimates for the inferred emissivity shape and scaled emissivity for both the physics-conditioned and unconditioned models. Results are shown for a dataset of 500 radiance, with 1280 MC-samples comprising   $\hat{p}(\eps|\rad)$ for each sample.}
	\label{Fig:LogLikelihood}
\end{figure}

We next seek to quantify the match of the inferred distribution to the target emissivity numerically. To do so, we consider a log-likelihood metric of the ground truth emissivity $\eps$ being a member of the inferred distribution $\hat{p}(\eps|\rad)$.
Taking the empirical mean and standard deviation of the MC distribution as $\mu_{\hat{p}}(\lambda)$ and $\sigma_{\hat{p}}(\lambda)$, where the argument $\lambda$ explicitly notes wavelength dependence,  the likelihood for a given emissivity spectrum being apart of the model's inferred distribution can be readily computed over each wavelength band of the LWIR spectrum.
The results of the likelihood estimation are summarized by Fig~\ref{Fig:LogLikelihood} for both normalized emissivity ($\tilde\eps$) and scaled emissivity ($\eps$), and for both physics-conditioned and unconditioned modeling schemes.
The solid lines and shaded regions of Fig~\ref{Fig:LogLikelihood} denote the average likelihood and standard deviation of the likelihood for 500 random samples of test data.

Both models display comparable performance with respect to emissivity shape, with the log-likelihood ranging between $-$1.5 (error of $\lesssim 2$ standard deviations) and 0.5 (error of $\lesssim 1$ standard deviations) indicating a very strong match between the target spectra and inferred distribution in general. 
There is a notable wavelength dependence on the likelihood scores, with the models struggling to infer the correct shape in the low-wavelength regions (between 7.5 and 9 $\mu$m). This is likely a result of the low transmission of radiance over these spectral bands, occluding most ground-leaving radiance for the low-wavelength regions (cf.~Fig~\ref{Fig:Lp}).
Moreover, the comparable performance between conditioned and unconditioned models re-affirms the empirically drawn conclusion that whitened radiance posses rich information regarding the spectral features of the underlying emissivity, with physics-based conditioning offering some (but not significant) improvement. We note this also aligns with the loss curves presented in Fig~\ref{Fig:Loss}.

Regarding the scaled emissivity of Fig~\ref{Fig:LogLikelihood}, higher log-likelihood scores and greater deviation between the conditioned and unconditioned modeling schemes is observed on average. Moreover, there is less wavelength dependent variation, which is expected due to the prominence of relatively low variance emissivity curves. For such signals, capturing the correct mean (or offset) of the emissivity curve is vital; incorporating the physics-based conditioning modules is thus highly beneficial for learning the relative amplitude and variance of the target emissivity curve for a given radiance sample. This too aligns with Fig~\ref{Fig:Loss}, where the physics-conditioned model out-performs the unconditioned model for tasks such as scaled emissivity reconstruction and propagation reconstruction.
Hence, while both models show good performance in terms of capturing the target spectra via $\hat{p}(\eps|\rad)$, a performance increase is recovered via physics-based conditioning.

\subsection{Material Identification}
\label{subsec:material_id}

With the end goal of most HSI modeling frameworks being the remote detection of materials, we conduct a final set of numerical experiments to evaluate the efficacy for material ID. 
To do so, the laboratory measured materials $\epsset$ were  propagated  into test HSI cubes via Eq~\eqref{eq:fp} without applying $\mathcal{GP}$ modulations. The generated empirical distributions were subsequently compared to the candidate spectra to determine the most likely candidates.
In what follows, the matching performance in a qualitative sense for a set of four common materials is presented first, and a quantitative analysis of the likelihood of returning the target material as a top-$K$ match for various values of $K$ is subsequently given.
\begin{figure*}
	\fullfigdouble{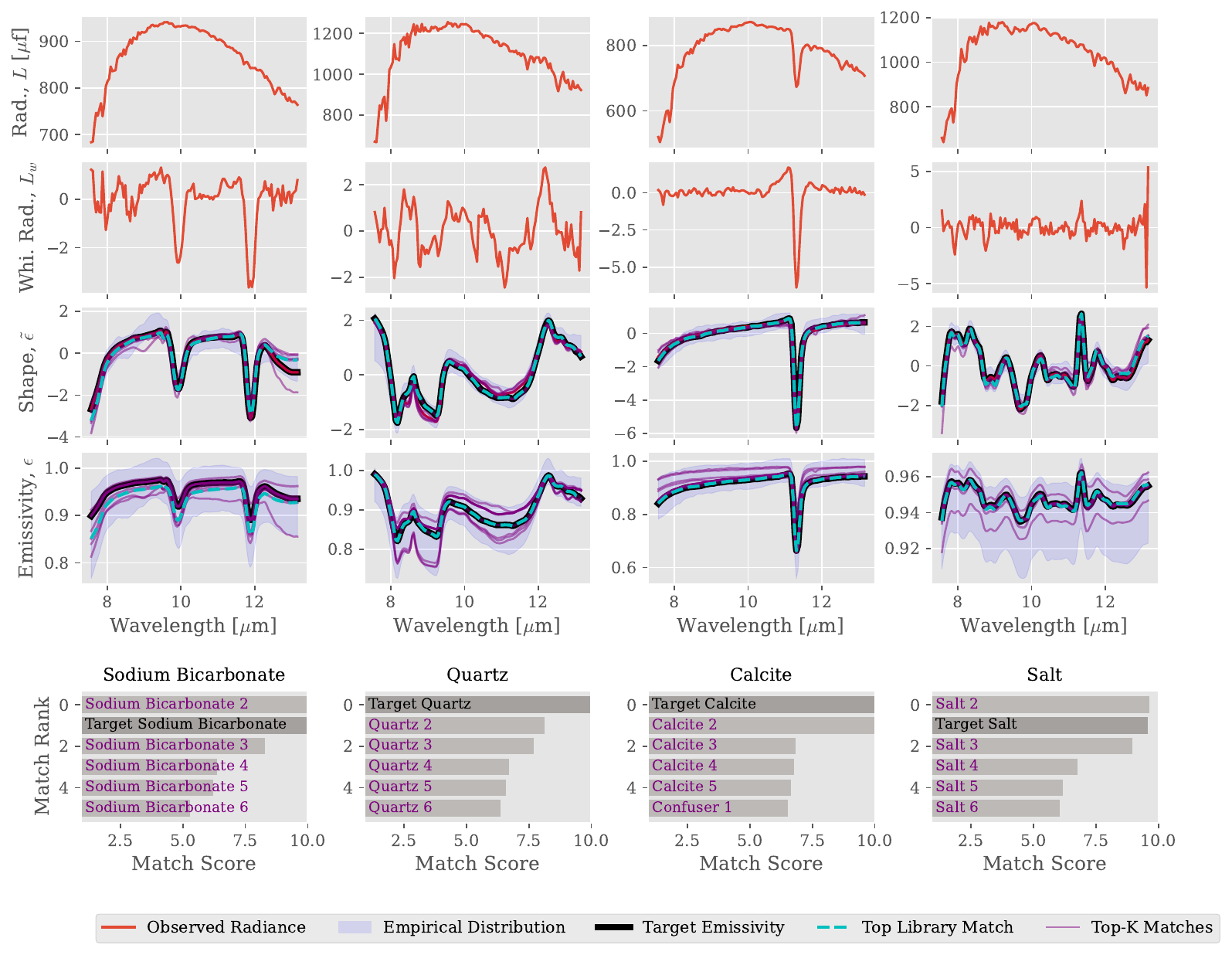}
	\caption{Probabilistic material ID. Top:  observed radiance values $\rad$ and whitened radiance $\rad_w$ for four sample materials propagated into an HSI scene. Middle:  the distribution of the estimated emissivity shape $\hat{p}({\tilde\eps}|\rad)$ and scaled emissivity $\eps$  in shaded blue with the true emissivity in black, top match in dashed cyan, and top-K matches in purple. Bottom: abbreviated material names from the library of the top-K matches along with their match score, with the  true target material appearing in black.
	}
	\label{Fig:MaterialID}
\end{figure*}

\subsubsection{Variance-Aware Framework for Material Matching}

The spectral matching process is as follows.
After producing $[\hat\eps]_N$ and $[\hat{\tilde\eps}]_N$ for the propagated radiance measures, the most-likely materials may be identified by quantifying how well each spectrum of  $\epsset$ aligns with the empirical distribution.
To draw such comparisons, we consider the 
Mahalanobis distance between the empirically generated set of $N$ spectral curves, $[\hat\eps]_N$, and each candidate spectra $\eps\in\epsset$, which we denote as $d_{\rm MD}(\hat{p},\eps)$. 
Defining $\bm{\mu}_{\hat{p}}$ and $\bm{\Sigma}_{\hat{p}}$ and the mean vector and covariance matrix of $[\hat\eps]_N$, respectively, the measure $d_{\rm MD}(\hat{p},\eps) $ is given as,
\begin{equation}
	d_{\rm MD}(\hat{p},\eps 
	) 
	=
	\sqrt{
		({\epsilon}-\bm{\mu}_{\hat{p}} )^{\intercal}\bm{\Sigma}_{\hat{p}}^{-1}
		({\epsilon}-\bm{\mu}_{\hat{p}} ).
	}
\end{equation}
With this definition, the match-score between an empirically generated distribution of normalized and scaled emissivity (denoted $\tilde{\hat{p}}$ and $\hat{p}$ herein) and a corresponding scaled and unscaled emissivity candidate from the spectral library is given as a combined measure of $d_{\rm MD}$ applied in both the normalized (shape-matching) and scaled (overall spectra matching) spaces,
\begin{equation}
		M(\hat{p},\eps) = \left[\underbrace{\left(d_{\text{MD}}(\tilde{\hat{p}},\tilde\eps) + d_{\text{MD}}({\hat{p}},\eps)\right)^{-1 }}_{M_0(\hat{p},\eps) }-\zeta\right]^{1/2} ,\\
\label{Eq:Match}
\end{equation} 
where the quantity $\zeta= \left\{\min M_0(\hat{p},\eps); \eps\in\epsset\right\}$ is a normalization  constant used so that 0 is the the least-likely score of the library. A key advantage of this metric is its ability to account for variance dependencies over wavelength thanks to the incorporation of the full covariance matrix, thus making it a variance-aware distribution-based metric.

The probabilistic material ID framework was applied to the propagated radiance's of measured samples of
baking soda, quartz, calcite, and salt.
These materials were selected for three primary reasons. First, They are common materials with relative prominence in everyday HSI scenes. Second, for each, there exists multiple similar measurements in the library (typically of the same material, but with different parameters, e.g., grain size, color, etc) allowing us to measure the model's ability of returning high-probability scores for  materials with similar spectra. Finally, the selected materials exhibit diversity in spectral features across the LWIR spectrum allowing for the demonstration of the model on four unique material classes.

The results of the material ID experiment are given by Fig~\ref{Fig:MaterialID}.
The empirically estimated distributions $\hat{p}(\eps|\rad)$ and $\hat{p}(\tilde\eps|\rad)$ (depcited with blue shading) are well-aligned with the spectral signature of the target material for each example.  
The top match (shown in dashed cyan) and ground truth (shown in black) are the same for two of the four examples, indicating the highest score of Eq~\eqref{Eq:Match} being designated to the correct material.
The top-6 matches are shown for each example, along with their scores from Eq~\eqref{Eq:Match}. 
For the samples where the top match did not align with the sample material, the identified top match is similar to the target spectrum in shape and scale.  Moreover, the top material predictions of Fig~\ref{Fig:MaterialID} are all of the same material class (i.e., quartz, salt,etc.) as the target spectrum.
The results show that the target material is returned as a top-5 match for each example, and that high probability is assigned to the target material even if it is not recovered as the top match. 
Moreover, when looking to the alternative materials that are selected as highly-probable, it is seen that they are all similar in nature to the target material, typically having the same composition but with varying measurement hyperparameters (such as particle size). 

The results of Fig~\ref{Fig:MaterialID} highlight several notable advantages of the probabilistic inverse modeling framework.
Whereas traditional classification-based target identification networks return a discrete label without a clear explanation of confidence or likely alternatives, the presented  model herein correctly returns high-probability spectra  and a measure of alignment for each material.
Moreover, for the examples of  Fig~\ref{Fig:MaterialID}, the nearly equally high scores are returned for materials of the same material class. 
This affirms the model's ability to model general spectral features of the inverse process in a regularized (probabilistic) framework, thus encouraging the mapping of more general material classes rather than fixating on data-specific features (as is common when training deterministic models on a fixed training set).
This introduced flexibility is a result of the enforced uncertainty of the latent distribution and the emissivity modulation scheme described in section~\ref{subsec:epsmod}. 
Hence, we conclude that the developed probabilistic inverse model also provides a meaningful tool for probabilistic material ID.

\subsubsection{Quantitative Comparison to Variance-Neglecting Frameworks}
As a last experiment, we evaluate the ``hit rate" of the model as a function of the number of candidate spectra considered, denoted $K$. 
In this context, a ``hit'' is defined as the model returning the target material within the top $K$ matches to the inferred emissivity distribution.
Moreover, we consider a pseudo-deterministic limit of our model by comparing the matching scheme of Eq~\eqref{Eq:Match} to two expectation-based measures. Denoting the expected value for $\hat{p}(\eps|L)$ over samples of the latent posterior as $\mathbb{E}_{z\sim q_\phi}\left[\hat{p}(\eps|L)\right]$, we can consider distance metrics to the empirical expectation to be a variance-neglecting matching scheme.
To this end, we consider the cosine distance (CD) and $L_2$ norm as matching criteria between elements of the candidate spectra set $\epsset$ and the expectation vectors,
\begin{align}
	d_{\rm CD}(\hat{p},\eps)
	 &= \text{CD}(\mathbb{E}_{z\sim q_\phi}\left[\hat{p}(\eps|L)\right],\epsilon)
	\label{EQ:det1}
	\\
	d_{\rm L2}(\hat{p},\eps)
	&= \left|\left|\mathbb{E}_{z\sim q_\phi}\left[\hat{p}(\eps|L)\right]-\epsilon\right|\right|_2.
	\label{EQ:det2}
\end{align}
Since the wavelength-dependent variability and variation correlations that are present in $\hat{p}(\eps|L)$  are not captured by $\mathbb{E}_{z\sim q_\phi}\left[\hat{p}(\eps|L)\right]$, the expectation-based criteria is considered variance-neglecting. 
Comparing the hit rate of the variance-aware Eq~\eqref{Eq:Match} to the expectation based Eqs~\eqref{EQ:det1} and \eqref{EQ:det2} provides a direct measure to test the benefit of the probabilistic modeling framework in the context of material ID.

\begin{figure}[t!] 
	\centering
	\begin{subfigure}{\subfigurewidth}
		\fullfigdouble{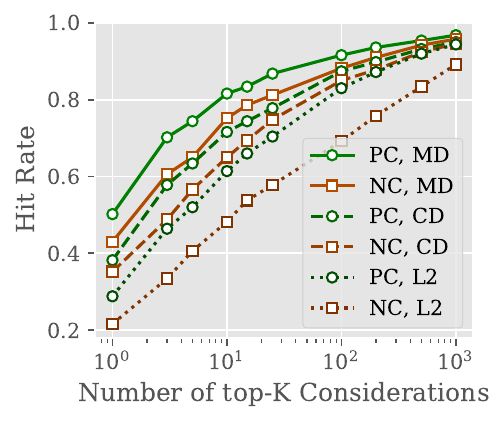}
		\caption{$\min(\alpha)=0.1$}
	\end{subfigure}%
	\begin{subfigure}{\subfigurewidth}
		\fullfigdouble{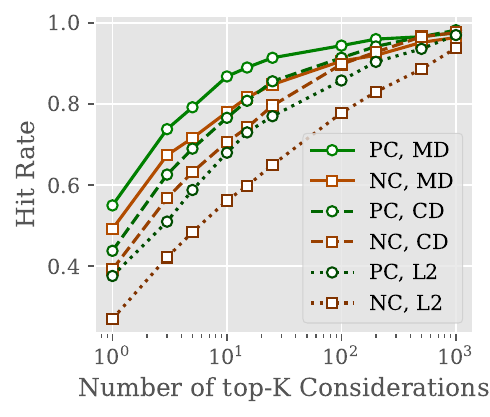}
		\caption{$\min(\alpha)=0.25$}
	\end{subfigure} 
	\begin{subfigure}{\subfigurewidth}
		\fullfigdouble{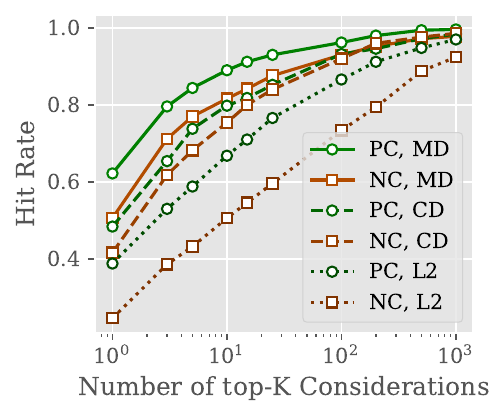}
		\caption{$\min(\alpha)=0.5$}
	\end{subfigure}%
	\begin{subfigure}{\subfigurewidth}
		\fullfigdouble{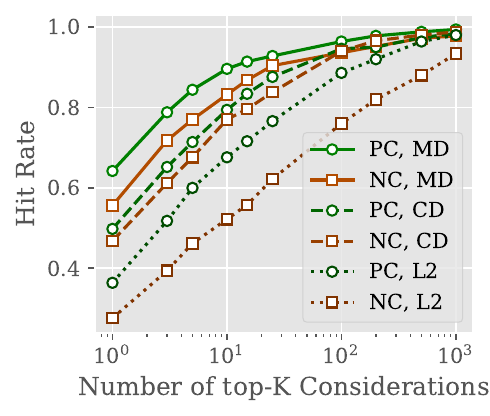}
		\caption{$\min(\alpha)=0.75$}
	\end{subfigure}
	\caption{The probability of including the target material spectrum in the top-$K$ matches (e.g., hit rate) for sample sets with minimum allowable strengths of (a) 0.1, (b) 0.25, (c) 0.5, and (d) 0.75.
		The six lines correspond to physics-conditioned (PC) vs non-conditioned (NC) models, and to distribution-based (MD),  expectation-based cosine distance (CD), and expectation-based L2 distance (L2) matching schemes.
	}
	\label{Fig:HitRate}
\end{figure}

The results of the hit-rate experiment are depicted by Fig~\ref{Fig:HitRate} for sub-trial data sets with minimum strengths of $\alpha=$ 0.1, 0.25, 0.5, and 0.75, and for $K\in[1,1000]$.
Curves showing hit-rate versus $K$ are given for both conditioned and unconditioned modeling schemes, and for the three proposed material matching frameworks.
Regarding the interpretation of the curves, the faster the convergence to 1, the better the performance.
In this light, the proposed variance-aware and physics-conditioned matching framework is the clear front-runner for all sub-trials.
Ranking second is the non-conditioned and variance-aware framework which, along with its conditioned counterpart, outperform all pseudo-deterministic matching frameworks; this indicates the benefit of incorporating structured uncertainty when identifying candidate materials. 
Moreover, the performances of all schemes improve as the minimum allowable strength grows which is an expected result as more of the target emissivity is present in the signal (as opposed to background) for higher $\alpha$ values.

Commenting on the overall performance of the physics-conditioned and variance aware results of Fig~\ref{Fig:HitRate}, it is shown that even for data sets containing low-$\alpha$ samples that the probability of identifying the specific target measurement is over 80\% when incorporating just the first 10 matches. Moreover, per Fig~\ref{Fig:MaterialID}, it is likely that the other 9 matches are highly similar to the target in spectral makeup. 
When weak signals are excluded (e.g., $\alpha>0.75$), this metric raises near 90\%. 
Hence, the results of Fig~\ref{Fig:HitRate} lead to the following conclusions: (i) that the probabilistic material ID framework is effective for return target materials with high success rates, (ii) that the physics conditioning results in a quantitatively more effective inverse model, and (iii) that awareness of the structured uncertainty over the wavelength domain is beneficial for performing robust matching with the inverse model, e.g., that the probabilistic nature of the model is beneficial for material ID.


\section{Conclusions and Suggested Future Work}
\label{SEC:Conclusions}

This work presented a deep-learning framework for probabilistic emissivity retrieval, e.g., a hyperspectral inverse model with quantifiable uncertainty.
The conditional distribution $p(\eps|\rad)$ was inferred based on whitened and un-whitened sensor measurements, for which an ELBO objective was derived. 
A latent-variable model was constructed to satisfy the training objective which utilized a combination of ResNet and FNO-MLP type encoding and decoding operations, with a normalizing flow applied to the latent posterior.
Furthermore, a physics-based conditioning mechanism was incorporated whereby the auxiliary parameters of the propagation equation dictating the mapping from $\eps$ to $\rad$ (e.g., $\atm$ and $\Lbg$ of Eq~\eqref{eq:fp}) were learned based on surrounding cube's radiance measurements, providing vital contextualization to a single-pixel's observation for the task of inverting Eq~\eqref{eq:fp}.
The efficacy of the proposed modeling strategy was demonstrated through several numerical experiments. It was shown that the MC estimates of $p(\eps|\rad)$ effectively capture target emissivity, and that the physics-based conditioning improves inference of $p(\eps|\rad)$ as compared to unconditioned modeling (cf.~Figs~\ref{Fig:Loss}, \ref{Fig:KDE}, and \ref{Fig:LogLikelihood}).
Moreover, we demonstrated that the proposed model may perform as a probabilistic material ID mechanism (cf.~Fig~\ref{Fig:MaterialID}), and that a physics-conditioned and probabilistic framework for material matching outperforms unconditioned and pseudo-deterministic schemes (cf.~Fig~\ref{Fig:HitRate}).

The aforementioned findings lead to the general conclusion that the proposed model is effective for performing targeting emissivity retrieval with quantified uncertainty.
With this come several notable advantages and implications.
First, the proposed framework provides a means for both quantifying the likely spectral variability in the LWIR range given a radiance measure and for determining likely material matches (in contrast to standard categorical classification models).
Second, the proposed deep learning framework is relatively cheap to query (in contrast to conventional atmospherics compensation codes~\cite{Manolakis2002}). Moreover, it's application extends beyond training distributions which was verified by separating testing and training cubes by acquisition date (a challenge for deep learning frameworks); this generality is thanks in part to the physics-based conditioning mechanisms.
Finally, in contrast to conventional emissivity retrieval algorithms which consider all ground leaving radiance of a pixel (e.g., mixed background and target), our proposed framework learns to decouple the background emissivity from the target spectrum (e.g., the portion of the pixel's emissivity that is not considered background), even when the relative target strength is low (on the order of 10\%).
Hence, this work provides valuable assets to analysts which can be leveraged in LWIR HSI sensing and exploitation efforts.

While this work has focused on synthesized data, e.g., data which was generated by propagating laboratory measured spectra (or perturbations of it) into measured HSI scenes via Eq~\eqref{eq:fp}, future work could further verify this modeling framework on real-world measurements (in contrast to the current work which considers laboratory-measured spectra synthesized into real-world HSI scenes). 
Of course, such a task is accompanied by many challenges, such as knowing the ground-leaving target emissivity and having the resource to perform the accompanying aerial measurements. 
Future work could also consider more intricate conditioning schemes, i.e., by overriding the uniform background and atmosphere assumption in lieu of localized conditioning (in contrast to cube-wide conditioning) for a given region of interest. 
Lastly, as this work focused exclusively on solids, future work could extend this framework to gas plume identification.	 
	\section*{Acknowledgments}
	Funding was provided by the Laboratory Directed Research and Development Office through project \#20240779PRD1. The authors acknowledge the Aerospace Corporation for collecting and providing the historical airborne LWIR data from
	the Los Angeles basin area.\\
	
	LA-UR-25-25141
	
	\bibliographystyle{elsarticle-num} 
	\bibliography{HSIbib}

\begin{thebibliography}{10}
\expandafter\ifx\csname url\endcsname\relax
  \def\url#1{\texttt{#1}}\fi
\expandafter\ifx\csname urlprefix\endcsname\relax\def\urlprefix{URL }\fi
\expandafter\ifx\csname href\endcsname\relax
  \def\href#1#2{#2} \def\path#1{#1}\fi

\bibitem{Manolakis2002}
D.~Manolakis, G.~Shaw, Detection algorithms for hyperspectral imaging
  applications, IEEE signal processing magazine 19~(1) (2002) 29--43.
\newblock \href {https://doi.org/10.1109/79.974724}
  {\path{doi:10.1109/79.974724}}.

\bibitem{Manolakis2019}
D.~Manolakis, M.~Pieper, E.~Truslow, R.~Lockwood, A.~Weisner, J.~Jacobson,
  T.~Cooley, Longwave infrared hyperspectral imaging: Principles, progress, and
  challenges, IEEE Geoscience and Remote Sensing Magazine 7~(2) (2019) 72--100.
\newblock \href {https://doi.org/10.1109/MGRS.2018.2889610}
  {\path{doi:10.1109/MGRS.2018.2889610}}.

\bibitem{Dale2013}
L.~M. Dale, A.~Thewis, C.~Boudry, I.~Rotar, P.~Dardenne, V.~Baeten, J.~A.~F.
  Pierna, Hyperspectral imaging applications in agriculture and agro-food
  product quality and safety control: A review, Applied Spectroscopy Reviews
  48~(2) (2013) 142--159.
\newblock \href {https://doi.org/10.1080/05704928.2012.705800}
  {\path{doi:10.1080/05704928.2012.705800}}.

\bibitem{Garcia2024}
Y.~E. Garc{\'\i}a-Vera, A.~Poloch{\`e}-Arango, C.~A. Mendivelso-Fajardo, F.~J.
  Guti{\'e}rrez-Bernal, Hyperspectral image analysis and machine learning
  techniques for crop disease detection and identification: A review,
  Sustainability 16~(14) (2024) 6064.
\newblock \href {https://doi.org/10.3390/su16146064}
  {\path{doi:10.3390/su16146064}}.

\bibitem{Sabins1999}
F.~F. Sabins, Remote sensing for mineral exploration, Ore geology reviews
  14~(3-4) (1999) 157--183.
\newblock \href {https://doi.org/10.1016/S0169-1368(99)00007-4}
  {\path{doi:10.1016/S0169-1368(99)00007-4}}.

\bibitem{heiden2012urban}
U.~Heiden, W.~Heldens, S.~Roessner, K.~Segl, T.~Esch, A.~Mueller, Urban
  structure type characterization using hyperspectral remote sensing and height
  information, Landscape and urban Planning 105~(4) (2012) 361--375.

\bibitem{ramdani2013urban}
F.~Ramdani, Urban vegetation mapping from fused hyperspectral image and lidar
  data with application to monitor urban tree heights, Journal of Geographic
  Information System 5~(4) (2013) 404--408.

\bibitem{zhang2020htd}
G.~Zhang, S.~Zhao, W.~Li, Q.~Du, Q.~Ran, R.~Tao, Htd-net: A deep convolutional
  neural network for target detection in hyperspectral imagery, Remote Sensing
  12~(9) (2020) 1489.

\bibitem{zhu2020two}
D.~Zhu, B.~Du, L.~Zhang, Two-stream convolutional networks for hyperspectral
  target detection, IEEE Transactions on Geoscience and Remote Sensing 59~(8)
  (2020) 6907--6921.

\bibitem{Boonmee2006}
M.~Boonmee, J.~R. Schott, D.~W. Messinger, Land surface temperature and
  emissivity retrieval from thermal infrared hyperspectral imagery, in:
  Algorithms and Technologies for Multispectral, Hyperspectral, and
  Ultraspectral Imagery XII, Vol. 6233, SPIE, 2006, pp. 678--688.

\bibitem{Li2007}
J.~Li, J.~Li, E.~Weisz, D.~K. Zhou, Physical retrieval of surface emissivity
  spectrum from hyperspectral infrared radiances, Geophysical Research Letters
  34~(16) (2007).
\newblock \href {https://doi.org/10.1029/2007GL030543}
  {\path{doi:10.1029/2007GL030543}}.

\bibitem{Xu2021}
F.~Xu, J.~Sun, G.~Cervone, M.~Salvador, Ill-posed surface emissivity retrieval
  from multi-geometry hyperspectral images using a hybrid deep neural network,
  arXiv preprint (2021).
\newblock \href {https://doi.org/arXiv:2107.04631}
  {\path{doi:arXiv:2107.04631}}.

\bibitem{Mcelhinney2022}
O.~McElhinney, M.~Pieper, D.~Manolakis, C.~Loughlin, V.~Ingle, R.~Bostick,
  A.~Weisner, Spline based emissivity retrieval for lwir hyperspectral imagery,
  in: Algorithms, Technologies, and Applications for Multispectral and
  Hyperspectral Imaging XXVIII, Vol. 12094, SPIE, 2022, pp. 226--236.

\bibitem{Sifnaios2024}
S.~Sifnaios, G.~Arvanitakis, F.~K. Konstantinidis, G.~Tsimiklis, A.~Amditis,
  P.~Frangos, A deep learning approach for pixel-level material classification
  via hyperspectral imaging, arXiv preprint arXiv:2409.13498 (2024).

\bibitem{Klein2023}
N.~Klein, A.~Carr, Z.~Hampel-Arias, A.~Zastrow, A.~Ziemann, E.~Flynn,
  Hyperspectral target identification using physics-guided neural networks with
  explainability and feature attribution, in: IGARSS 2023 - 2023 IEEE
  International Geoscience and Remote Sensing Symposium, 2023, pp. 946--949.
\newblock \href {https://doi.org/10.1109/IGARSS47720.2023.10250812}
  {\path{doi:10.1109/IGARSS47720.2023.10250812}}.

\bibitem{Buckland2017}
K.~N. Buckland, S.~J. Young, E.~R. Keim, B.~R. Johnson, P.~D. Johnson, D.~M.
  Tratt, Tracking and quantification of gaseous chemical plumes from
  anthropogenic emission sources within the los angeles basin, Remote Sensing
  of Environment 201 (2017) 275--296.

\bibitem{Distasio2010}
R.~J. DiStasio~Jr, R.~G. Resmini, Atmospheric compensation of thermal infrared
  hyperspectral imagery with the emissive empirical line method and the
  in-scene atmospheric compensation algorithms: a comparison, in: Algorithms
  and technologies for multispectral, hyperspectral, and ultraspectral imagery
  XVI, Vol. 7695, SPIE, 2010, pp. 805--816.
\newblock \href {https://doi.org/10.1117/12.849898}
  {\path{doi:10.1117/12.849898}}.

\bibitem{Khan2018}
M.~J. Khan, H.~S. Khan, A.~Yousaf, K.~Khurshid, A.~Abbas, Modern trends in
  hyperspectral image analysis: A review, Ieee Access 6 (2018) 14118--14129.
\newblock \href {https://doi.org/10.1109/ACCESS.2018.2812999}
  {\path{doi:10.1109/ACCESS.2018.2812999}}.

\bibitem{Acito2019}
N.~Acito, M.~Diani, G.~Corsini, Coupled subspace-based atmospheric compensation
  of lwir hyperspectral data, IEEE Transactions on Geoscience and Remote
  Sensing 57~(8) (2019) 5224--5238.

\bibitem{Signoroni2019}
A.~Signoroni, M.~Savardi, A.~Baronio, S.~Benini, Deep learning meets
  hyperspectral image analysis: A multidisciplinary review, Journal of Imaging
  5~(5) (2019).
\newblock \href {https://doi.org/10.3390/jimaging5050052}
  {\path{doi:10.3390/jimaging5050052}}.

\bibitem{Rasti2020}
B.~Rasti, D.~Hong, R.~Hang, P.~Ghamisi, X.~Kang, J.~Chanussot, J.~A.
  Benediktsson, Feature extraction for hyperspectral imagery: The evolution
  from shallow to deep: Overview and toolbox, IEEE Geoscience and Remote
  Sensing Magazine 8~(4) (2020) 60--88.
\newblock \href {https://doi.org/10.1109/MGRS.2020.2979764}
  {\path{doi:10.1109/MGRS.2020.2979764}}.

\bibitem{Gupta2022}
K.~Gupta, V.~Bajaj, I.~A. Ansari, An improved deep learning model for automated
  detection of bbb using s-t spectrograms of smoothed vep signal, IEEE Sensors
  Journal 22~(9) (2022) 8830--8837.
\newblock \href {https://doi.org/10.1109/JSEN.2022.3162022}
  {\path{doi:10.1109/JSEN.2022.3162022}}.

\bibitem{Xue2021}
Z.~Xue, X.~Yu, X.~Tan, B.~Liu, A.~Yu, X.~Wei, Multiscale deep learning network
  with self-calibrated convolution for hyperspectral and lidar data
  collaborative classification, IEEE Transactions on Geoscience and Remote
  Sensing 60 (2021) 1--16.
\newblock \href {https://doi.org/10.1109/TGRS.2021.3106025}
  {\path{doi:10.1109/TGRS.2021.3106025}}.

\bibitem{Tan2024}
P.~Tan, J.~Li, G.~Zhang, L.~Zhao, Multi-scale and multi-modal contrastive
  learning for hyperspectral and lidar classification, in: 2024 2nd
  International Conference on Computer, Vision and Intelligent Technology
  (ICCVIT), IEEE, 2024, pp. 1--6.
\newblock \href {https://doi.org/10.1109/ICCVIT63928.2024.10872424}
  {\path{doi:10.1109/ICCVIT63928.2024.10872424}}.

\bibitem{Wang2021}
C.~Wang, B.~Liu, L.~Liu, Y.~Zhu, J.~Hou, P.~Liu, X.~Li, A review of deep
  learning used in the hyperspectral image analysis for agriculture, Artificial
  Intelligence Review 54~(7) (2021) 5205--5253.

\bibitem{Klein2023a}
N.~Klein, A.~Carr, Z.~Hampel-Arias, A.~Zastrow, A.~Ziemann, E.~Flynn,
  Hyperspectral target identification using physics-guided neural networks with
  explainability and feature attribution, in: IGARSS 2023-2023 IEEE
  International Geoscience and Remote Sensing Symposium, IEEE, 2023, pp.
  946--949.
\newblock \href {https://doi.org/10.1109/IGARSS52108.2023.10283350}
  {\path{doi:10.1109/IGARSS52108.2023.10283350}}.

\bibitem{Guanter2007}
L.~Guanter, M.~Del Carmen Gonz{\'a}lez-Sanpedro, J.~Moreno, A method for the
  atmospheric correction of envisat/meris data over land targets, International
  Journal of Remote Sensing 28~(3-4) (2007) 709--728.
\newblock \href {https://doi.org/10.1080/01431160600815525}
  {\path{doi:10.1080/01431160600815525}}.

\bibitem{Thompson2016}
D.~R. Thompson, D.~A. Roberts, B.~C. Gao, R.~O. Green, L.~Guild, K.~Hayashi,
  R.~Kudela, S.~Palacios, Atmospheric correction with the bayesian empirical
  line, Optics express 24~(3) (2016) 2134--2144.

\bibitem{Patel2020}
A.~K. Patel, J.~K. Ghosh, Quantitative analysis of mixed pixels in
  hyperspectral image using fractal dimension technique, Journal of the Indian
  Society of Remote Sensing 48 (2020) 1237--1244.

\bibitem{He2023}
X.~He, Y.~Chen, L.~Huang, Bayesian deep learning for hyperspectral image
  classification with low uncertainty, IEEE Transactions on Geoscience and
  Remote Sensing 61 (2023) 1--16.
\newblock \href {https://doi.org/TGRS.2023.3257865}
  {\path{doi:TGRS.2023.3257865}}.

\bibitem{Seydgar2022}
M.~Seydgar, S.~Rahnamayan, P.~Ghamisi, A.~A. Bidgoli, Semisupervised
  hyperspectral image classification using a probabilistic pseudo-label
  generation framework, IEEE Transactions on Geoscience and Remote Sensing 60
  (2022) 1--18.
\newblock \href {https://doi.org/10.1109/TGRS.2022.3195924}
  {\path{doi:10.1109/TGRS.2022.3195924}}.

\bibitem{Yao2023}
D.~Yao, Z.~Zhi-li, Z.~Xiao-feng, C.~Wei, H.~Fang, C.~Yao-ming, W.-W. Cai, Deep
  hybrid: multi-graph neural network collaboration for hyperspectral image
  classification, Defence Technology 23 (2023) 164--176.
\newblock \href {https://doi.org/10.1016/j.dt.2022.02.007}
  {\path{doi:10.1016/j.dt.2022.02.007}}.

\bibitem{Ding2024}
Y.~Ding, Z.~Zhang, H.~Hu, F.~He, S.~Cheng, Y.~Zhang, Graph sample and
  aggregate-attention network for hyperspectral image classification, in: Graph
  Neural Network for Feature Extraction and Classification of Hyperspectral
  Remote Sensing Images, Springer, 2024, pp. 29--41.
\newblock \href {https://doi.org/10.1007/978-981-97-8009-9}
  {\path{doi:10.1007/978-981-97-8009-9}}.

\bibitem{Wu2024}
G.~Wu, M.~A. Al-qaness, D.~Al-Alimi, A.~Dahou, M.~Abd~Elaziz, A.~A. Ewees,
  Hyperspectral image classification using graph convolutional network: A
  comprehensive review, Expert Systems with Applications 257 (2024) 125106.
\newblock \href {https://doi.org/10.1016/j.eswa.2024.125106}
  {\path{doi:10.1016/j.eswa.2024.125106}}.

\bibitem{Manolakis2016}
D.~G. Manolakis, R.~B. Lockwood, T.~W. Cooley,
  \href{https://www.cambridge.org/core/books/hyperspectral-imaging-remote-sensing/E19050621F91BFCAF904CD1A8CB35F29}{Hyperspectral
  Imaging Remote Sensing: Physics, Sensors, and Algorithms}, Cambridge
  University Press, Cambridge, UK, 2016.
\newline\urlprefix\url{https://www.cambridge.org/core/books/hyperspectral-imaging-remote-sensing/E19050621F91BFCAF904CD1A8CB35F29}

\bibitem{Gao2006}
B.-C. Gao, C.~Davis, A.~Goetz, A review of atmospheric correction techniques
  for hyperspectral remote sensing of land surfaces and ocean color, in: 2006
  IEEE International Symposium on Geoscience and Remote Sensing, IEEE, 2006,
  pp. 1979--1981.

\bibitem{Adler2014}
S.~Adler-Golden, P.~Conforti, M.~Gagnon, P.~Tremblay, M.~Chamberland, Long-wave
  infrared surface reflectance spectra retrieved from telops hyper-cam imagery,
  in: Algorithms and Technologies for Multispectral, Hyperspectral, and
  Ultraspectral Imagery XX, Vol. 9088, SPIE, 2014, pp. 247--254.

\bibitem{Young2002}
S.~J. Young, B.~R. Johnson, J.~A. Hackwell, An in-scene method for atmospheric
  compensation of thermal hyperspectral data, Journal of Geophysical Research:
  Atmospheres 107~(D24) (2002) ACH--14.

\bibitem{Rahm2023}
M.~Rahm, F.~Kullander, M.~Bj{\"o}rck, L.~Sj{\"o}qvist, Turbulence and
  transmission effects on laser beam propagation in the swir and lwir bands,
  in: Environmental Effects on Light Propagation and Adaptive Systems VI, Vol.
  12731, SPIE, 2023, pp. 128--139.

\bibitem{Dinh2016}
L.~Dinh, J.~Sohl-Dickstein, S.~Bengio, Density estimation using real nvp, arXiv
  preprint arXiv:1605.08803 (2016).

\bibitem{Li2020}
Z.~Li, N.~Kovachki, K.~Azizzadenesheli, B.~Liu, K.~Bhattacharya, A.~Stuart,
  A.~Anandkumar, Fourier neural operator for parametric partial differential
  equations, arxiv, arXiv preprint arXiv:2010.08895 (2020).

\bibitem{Kovachki2023}
N.~Kovachki, Z.~Li, B.~Liu, K.~Azizzadenesheli, K.~Bhattacharya, A.~Stuart,
  A.~Anandkumar, Neural operator: Learning maps between function spaces with
  applications to pdes, Journal of Machine Learning Research 24~(89) (2023)
  1--97.

\bibitem{brochu2010tutorial}
E.~Brochu, V.~M. Cora, N.~De~Freitas, A tutorial on bayesian optimization of
  expensive cost functions, with application to active user modeling and
  hierarchical reinforcement learning, arXiv preprint arXiv:1012.2599 (2010).

\end{thebibliography}
	
	\clearpage
	\newpage 
	\appendices 
	 
\section{Evaluation of the ELBO loss}
\label{Apx:ELBO}
Here we provide the complete details of the derivation and numerical implementation of Eq~\eqref{EQ:ELBO} (or equivilently Eq~\eqref{EQ:ELBO_c}), along with details regarding its implementation.

\subsection{Derivation}
We begin with Eq~\eqref{EQ: log_liklihood_p(y|x)} and introduce a variational distribution $q_\phi(\z|\rad)$ to make the integration tractable,
\begin{equation}
	\begin{aligned}
	\LL  	&= \log \int\frac{q(\z|\rad)} {q(\z|\rad)}p(\z|\rad)p(\eps|z){\rm d}z\\
			&= \log\mathbb{E}_{z\sim q(\z|\rad)}\left[
			\frac{p(\z|\rad)p(\eps|z)}{q(\z|\rad)}\right] \\
			&\geq \mathbb{E}_{z\sim q(\z|\rad)}\left[\log
			\frac{p(\z|\rad)p(\eps|z)}{q(\z|\rad)}\right] 
	\end{aligned}
	\label{EQa1}
\end{equation}
which is the un-factorized ELBO objective.
We may expand Eq~\eqref{EQa1} as,
\begin{equation}
	\ELBO = \mathbb{E}_{z\sim q(\z|\rad)}\left[
	\log p(\z|\rad) + \log p(\eps|\z) - \log q(\z|\rad)
	\right].
\end{equation}
Noting  $p(\z|\rad) =  {p(\rad|\z)p(z)}/{p(L)}$ by Bayes theorem, this turns to, 
\begin{equation}
		\log	p(\z|\rad)   = \log p(\rad|\z) + \log p(z) - \log p(L),
\end{equation} 
thus furnishing the objective
\begin{equation}
	\begin{aligned}
			\ELBO = \mathbb{E}_{z\sim q(\z|\rad)}\Big[&
			\log p(\rad|\z) + \log p(z) 
			\\
			&
			- \underbrace{\log p(L)}_{\text{constant}}  
			+ \log p(\eps|\z) - \log q(\z|\rad)
			\Big].
			\end{aligned}
\end{equation}
Since $p(\rad)$ is constant within the expectation computed over samples of $z\sim q(\z|\rad)$, it may be omitted from the objective giving,
\begin{equation}
	\begin{aligned}
	\ELBO = & \mathbb{E}_{z\sim q(\z|\rad)}\Big[
	\log p(\rad|\z)+\log p(\eps|\z) 	\Big]
	\\
	&+\underbrace{\mathbb{E}_{z\sim q(\z|\rad)}\Big[ \log p(z)   - \log q(\z|\rad)
	\Big]}_{-\kld\left( q(\z|\rad)||p(\z) \right)},
\end{aligned}
\end{equation}
which is the final form reported by Eq~\eqref{EQ:ELBO}.

\subsection{Implementation}

We utilize amortized inference and assume variational model forms,
\begin{equation}
	\begin{aligned}
			q_\phi(\z|\rad) 	\coloneqq	\mathcal{N}(\mu_\phi,\sigma_\phi^2), 		\ \ \ \						\{\mu_\phi,\sigma_\phi^2\}=f_\phi(\radset)  \\
		p_\theta(\rad|\z) 	\coloneqq		\mathcal{N}(\mu_\theta,\sigma_\theta^2), 		\ \ \ \					\{\mu_\theta,\sigma_\theta^2\}=f_\theta(\rad) \\
		p_\gamma(\eps|\z) 	\coloneqq	 	\mathcal{N}(\mu_\gamma,\sigma_\gamma^2), 		\ \ \ \					\{\mu_\gamma,\sigma_\gamma^2\}=f_\gamma(\rad) \\
	\end{aligned}
\end{equation}
We assume diagonal (but not constant) variances for all distributions, e.g., heteroscedastic Gaussian.
With this, the evaluation of Eq~\eqref{EQ:ELBO} may be numerically implemented as:
\begin{align}
	\mathbb{E}_{z\sim q_{\phi}(\z|\rad)} \log p_\theta(\rad|\z) &= \mathbb{E}_{z\sim q_\phi(z|x)} \left[\log \frac{1}{\sqrt{2\pi\sigma_\theta^2}} - \frac{||\rad^i-\mu_\theta||^2}{2\sigma_\theta^2} \right]\\
	\mathbb{E}_{z\sim q_{\phi}(\z|\rad)} \log p_\gamma(\eps|\z) &= \mathbb{E}_{z\sim q_\phi(z|x)} \left[\log \frac{1}{\sqrt{2\pi\sigma_\gamma^2}} - \frac{||\eps^i-\mu_\gamma||^2}{2\sigma_\gamma^2} \right]\\
	\kld\left( q(\z|\rad)||p(\z) \right) &= -\frac{1}{2}\left(1-\log\sigma_\phi^2-\mu_\phi^2-\sigma_\phi^2\right)
\end{align}

\section{Normalizing Flow}
\label{Apx:Flow}

The normalizing flow applied a series of invertibile functions to a sample of the base distribution $z_0\sim q_\phi^0(\z_0|\rad)$. In this work, the realNVP flow model presented in Ref~\cite{Dinh2016} is applied based on its scalability, expressivity, and efficient Jacobian calculations in comparison to competing flow models.
Briefly summarizing its structure, the flow model operates on samples of the base distribution by partitioning the latent sample $z_0$ into $z_{0a}$ and $z_{0b}$, and applying then computing their updated values as,
\begin{equation}
	\begin{aligned}
		z_{1b} &= z_{0a}\\
		z_{1b} &= z_{0b}\odot \exp(s^{(1)}(z_{0a})) + t^{(1)}(z_{0a}),
	\end{aligned}
\end{equation}
where $s^{(k)}(\square)$ and $t^{(k)}(\square)$ are scale and translation functions which are implemented as trainable MLP networks. 
For each successive layer of the flow, the partitioning order is switched so that the first or second halves of $z_k$ is designated as $z_{ka}$ and $z_{kb}$ in alternating fashion. 
The RealNVP is invertible with a block-diagonal Jacobian,
\begin{equation}
	\frac{\partial f_k}{\partial z_{k-1}} =\begin{bmatrix}
		\textbf{I} & \textbf{0}\\
		 \frac{\partial z_{kb} }{\partial z_{(k-1)a}}& \textrm{diag}(\exp(s(z_{(k-1)a})))
	\end{bmatrix}
\end{equation}
Thanks to the block-diagonal structure, the transformed density of the posterior may be conveniently expressed as,
\begin{equation}
		\log q_\phi^K(\z|\rad) 
			= \log q^0_\phi(\z_0|\rad) - \sum_{k=1}^{K}\sum_{i=1}^{N_k}s_i^{(k)}(z_{ka})
\end{equation}
where $N_k$ is the length of $s^{({k})}$'s output. Accordingly, the KL-divergence of the transformed posterior takes the form, 
\begin{equation}
	\begin{aligned}
		\kld[q^k(z|\rad)|p(z)] 
		 =& \kld[q_\phi^0(\z_0|\rad),p(\rad)] \\&-\mathbb{E}_{z_0 \sim q_\phi^0(z_0|x)} \sum_{k=1}^{K}\sum_{i=1}^{N_k}s_i^{(k)}(z_{ka})
	\end{aligned}
\end{equation}

\section{Atmosphere Reconstruction}
\begin{figure*}[h]\centering
	\includegraphics[width=.975\linewidth]{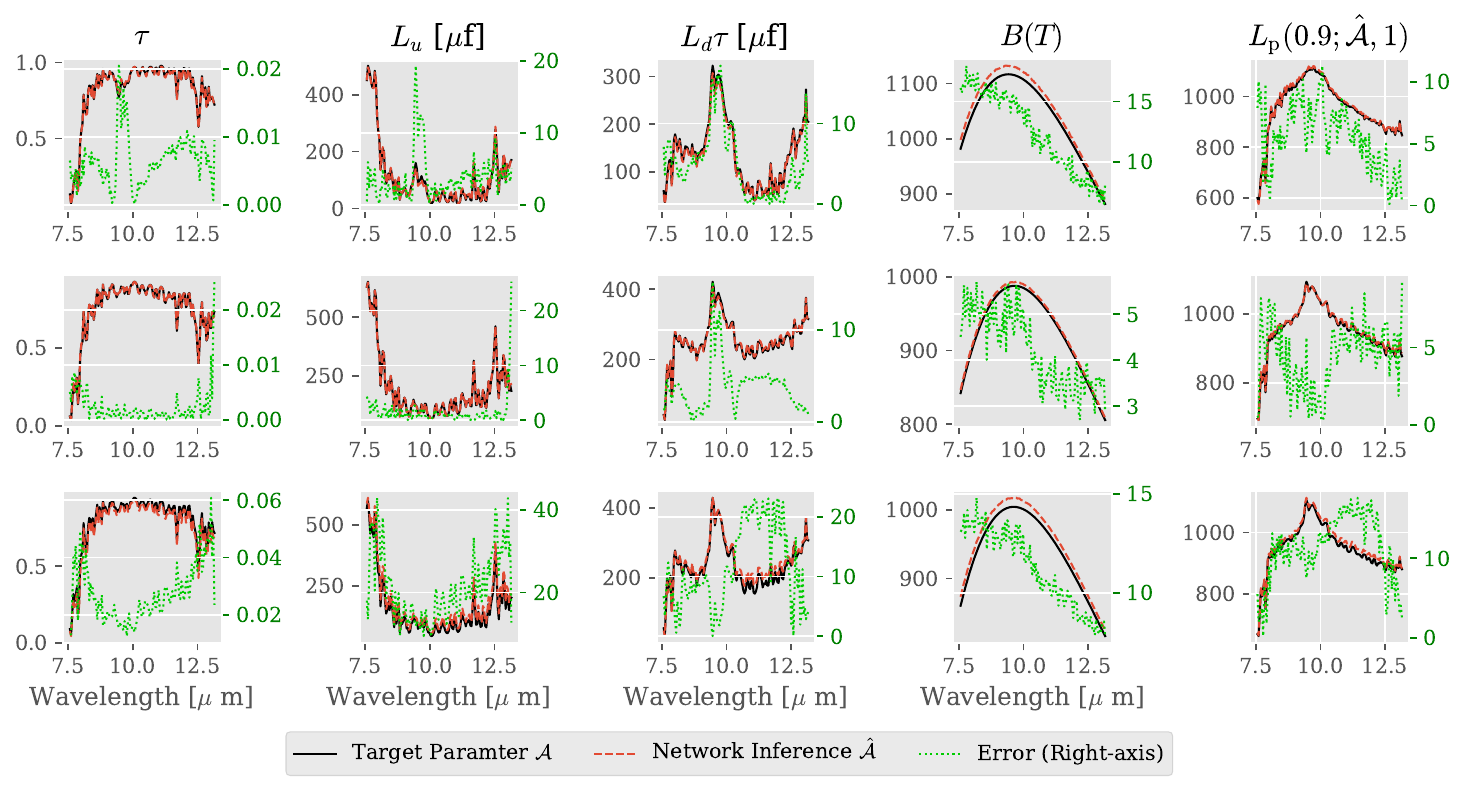}
	\caption{The propagation parameter estimates for three HSI cubes. Each row depicts a different cube, while the columns depict the transmission $\tau$, upwelling $L_u$, downwelling-transmission product $L_d\tau$, and black-body radiance $B$. The far-right column depicts the result of Eq~\eqref{eq:fp} for true and predicted parameters for a constant emissivity of 0.9 and unit strength. The error for each component is depicted in green and corresponds to the right axis of each subplot in scale.} 
	\label{Fig:Atms}
\end{figure*}

\label{Appendix:Atm}
Here we provide brief details of the reconstruction capacity for \atmnet given samples from an HSI scene. Fig~\ref{Fig:Atms} depicts the atmospheric parameters estimated from conventional physics-based regression codes as well as those inferred from \atmnet. 
Each row of Fig~\ref{Fig:Atms} depicts the results for a different HSI cube.
We note a good general agreement between the two for the three atmosphere's considered herein, and comment that a similar level of agreement is achieved by all cubes in general.
The specific cubes and atmopshereic selected for display by Fig~\ref{Fig:Atms} were chosen randomly.

%
%
%
	
\end{document}